\documentclass[lettersize,journal]{IEEEtran}
\usepackage{amsmath,amsfonts}
\usepackage{algorithmic}
\usepackage{algorithm}
\usepackage{array}
\usepackage[caption=false,font=normalsize,labelfont=sf,textfont=sf]{subfig}
\usepackage{textcomp}
\usepackage{stfloats}
\usepackage{hyperref}
\usepackage{url}
\usepackage{verbatim}
\usepackage{graphicx}
\newcommand{\norm}[1]{\left\lVert#1\right\rVert}

\begin{document}

\title{Cluster Specific Representation Learning}

\author{Mahalakshmi Sabanayagam, Omar Al-Dabooni, Pascal Esser
\thanks{This paper was produced by the Theoretical Foundations of Artificial Intelligence Group at the Technical University of Munich, Germany.}
\thanks{Correspondence email: \texttt{sabanaya@in.tum.de}.}}



\maketitle

\begin{abstract}
Representation learning aims to extract meaningful lower-dimensional embeddings from data, known as representations. Despite its widespread application, there is no established definition of a ``good'' representation. Typically, the representation quality is evaluated based on its performance in downstream tasks such as clustering, de-noising, etc. However, this task-specific approach has a limitation where a representation that performs well for one task may not necessarily be effective for another. This highlights the need for a more agnostic formulation, which is the focus of our work. We propose a downstream-agnostic formulation: \emph{when inherent clusters exist in the data, the representations should be specific to each cluster}. Under this idea, we develop a meta-algorithm that jointly learns cluster-specific representations and cluster assignments. As our approach is easy to integrate with any representation learning framework, we demonstrate its effectiveness in various setups, including Autoencoders, Variational Autoencoders, Contrastive learning models, and Restricted Boltzmann Machines. We qualitatively compare our cluster-specific embeddings to standard embeddings and downstream tasks such as de-noising and clustering. While our method slightly increases runtime and parameters compared to the standard model, the experiments clearly show that it extracts the inherent cluster structures in the data, resulting in improved performance in relevant applications.
\end{abstract}

\begin{IEEEkeywords}
Clustering algorithms, Representation learning, Unsupervised learning, Self-supervised learning
\end{IEEEkeywords}

\section{Introduction}



\begin{figure}
    \centering
    \includegraphics[width=0.9\linewidth]{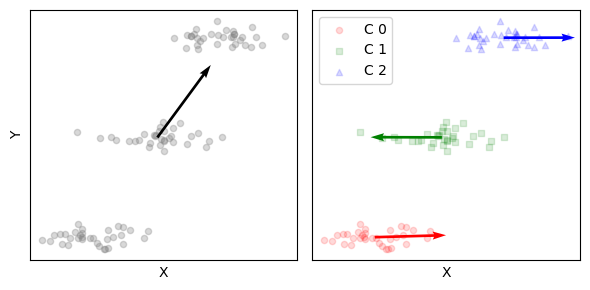}
    \caption{Representations obtained by a linear AE.
    \textbf{Left:} full dataset and representation of the full dataset (illustrated by the arrow).
    \textbf{Right:} cluster specific representations.
    }
    \label{fig:simpsons}
\end{figure}

Representation learning is a central paradigm in modern machine learning with the goal to obtain embeddings of the data, that can improve different downstream tasks such as classification or clustering.
However, evaluating the embedding with regard to downstream tasks restricts the transferability and generality of the considered embedding method.
We are therefore interested in a more \emph{intrinsic} definition of what constitutes a ``good'' embedding aka representation. 
While there are several ways such a notion could be defined, for example, with regards to removing noise \cite{vincent2008extracting,zhang2017beyond} or recovering topological structures \cite{carlsson2009topology,nigmetov2020topology}, we consider a \emph{cluster specific viewpoint - more specifically the obtained embedding should be specific for each cluster.} For an illustration, consider the example of three clusters shown in Figure~\ref{fig:simpsons}: we firstly note that linear Autoencoders (AEs) learn principal directions \cite{Kramer1991AIChE_AE} of the full data-set. Therefore, in Figure~\ref{fig:simpsons} (left), the representation, illustrated by the black arrow, gives the trend of the whole data-set but does not capture the structure in each cluster. On the other hand, Figure~\ref{fig:simpsons} (right) shows that if the embedding is cluster specific, the trends \emph{within the clusters} are captured well. 
Note that as the cluster specific embeddings capture the true intrinsic structure of the data, it is expected to perform well irrespective of the downstream task.
This data setting is also know as the \emph{simpson's paradox} \cite{Simpson1951stat} where the trend in clusters does not align with the trend that appears in the full dataset. Such settings are often observed in social science and medical science statistics \cite{Clifford1982,Holt2016}. 

The idea of learning cluster specific representations was proposed in \cite{esser2023improved} in the context of Autoencoders \cite{Kramer1991AIChE_AE} where the approach is to \emph{jointly learn the clustering together with the cluster specific encoder and decoder.} 
While \cite{esser2023improved} shows promising empirical results, it has two significant limitations that renders the technique impractical: firstly, the time and space model complexity scales with the number of clusters, making the approach not scalable, and secondly, the analysis is restricted to simple AEs. In this work, we improve on both aspects. 
%

\textbf{Contributions.}
More specifically we summarize the main contributions of this work as follows:
\begin{enumerate}
    \item We improve on the idea of cluster specific AEs \cite{esser2023improved} by proposing a more efficient implementation by only making a part of the embedding function cluster specific, significantly improving the time and model complexity.
    \item We show that the overall idea can be extended beyond AEs to  Variational AEs (VAEs) \cite{kingma2013auto}, Contrastive losses (CL), and Restricted Boltzmann machines (RBM) \cite{smolensky1986information}.
    \item Numerically, we show the performance on clustering and de-noising tasks can be improved through cluster specific embeddings. In addition, we analyze the cluster specific latent spaces.
\end{enumerate}

\textbf{Code.} The code to reproduce all numerical simulations, including hyperparameter search, is provided in the link  \url{https://tinyurl.com/2cshzdsy}.

\textbf{Notation}
We denote vectors as lowercase, $a$ and matrices as uppercase, $A$. Let $\norm{~\cdot~}^2$ be the standard square norm and a function $f$ parameterized by $\Theta$ as $f_\Theta$. Under abuse of notation we denote  $\Theta_j$ the parameters of function $f$ corresponding to cluster $j$ and $\{\Theta\}_1^k$ as the set of parameters $\{\Theta_1,\dots,\Theta_k\}$
Furthermore, let $1_n$ be the all ones vector of size $n$ and $\mathbb{I}_n$ the $n\times n$ identity matrix.

\section{Cluster Specific Representation Learning}
We now formalize the idea of learning cluster specific embedding functions as well as the optimization process and inference of new datapoints. This general idea can be applied to any embedding functions, and we subsequently outline how this is achieved for AEs, VAEs, CLs, and RBMs.

\textbf{Model definition.} Our overall goal is to embed  data-points $\{x_1,\dots x_n\},~x_i\in\mathbb{R}^d$ into an $h$-dimensional latent space, with $h\ll d$. Furthermore, we assume the data has $k$ clusters.
Before considering the cluster specific setting, let us recall the standard setting where $\mathcal{L}(\cdot)$ is loss function on the embedding obtained by the \emph{encoder function}, $g_\Psi:\mathbb{R}^d\rightarrow\mathbb{R}^h$, parameterized by $\Psi$. Note that while we outline the idea here for a loss function on the embedding, the general idea directly extends to reconstruction based losses such as AEs and VAEs. We optimize the parameters over the dataset:
\begin{align*}
    \min_{\Psi}\frac{1}{n} \sum_{i=1}^n\mathcal{L}\left(g_\Psi(x_i)\right).
\end{align*}
From this general form, we now define a cluster specific or \emph{``tensorized''\footnote{The notion of \emph{``tensorization''} arises from the fact that the stacked parameter matrices that are used to compute the embedding build a tensor in the cluster specific architectures.}} version, which allows for learning the clusters and cluster specific embeddings jointly.
We, therefore, first define a $k \times n$ matrix $S$, such that $S_{j,i}$ is the probability that data point $i$ belongs to cluster $j$. To ensure that it is a valid probability distribution, constraint $1_k^TS = 1_n^T,\ S_{j,i} \geq 0$ is added to the objective.
Furthermore, we define $k$ embedding functions $g_{\Psi_j},~j\in \{1,\ldots,k\}$, each specific to one cluster.
Combining the two ideas, we therefore define the cluster specific (tensorized) optimization $\mathbf{T}$  as: 
\begin{align*}
    \mathbf{T}: &\min_{\{\Psi\}_{1}^k,S}\frac{1}{n} \sum_{j=1}^n \sum_{i=1}^k S_{j,i}\left[\mathcal{L}\left(g_{\Psi_j}(x_i)\right)\right]\\
    &\text{s.t. ~ } 1_k^TS = 1_n^T,\ S_{j,i} \geq 0.
\end{align*}
While this formulation is straightforward, observe that summing over $k$ scales the number of parameters of the new model linearly with the number of considered clusters, making it computationally difficult for large data-sets or complex models. To circumvent this scalability issue, we now propose to split up the encoding function by considering firstly a \emph{joint encoding for all clusters} $g_{\Omega}:\mathbb{R}^d\rightarrow\mathbb{R}^{l}$ and subsequently a \emph{cluster specific encoding} $g_{\Psi_j}:\mathbb{R}^{l}\rightarrow\mathbb{R}^h$.
Therefore we propose the \emph{Partial Tensorized} optimization $\mathbf{PT}$ as
\begin{align}\label{eq: PT General}
    \mathbf{PT}: &\min_{\{\Psi\}_{1}^k,\Omega,S}\frac{1}{n} \sum_{j=1}^n \sum_{i=1}^k S_{j,i}\left[\mathcal{L}\left(g_{\Psi_j}\left(g_\Omega(x_i)\right)\right)\right]\\
    &\text{s.t. ~ } 1_k^TS = 1_n^T,\ S_{j,i} \geq 0.\nonumber
\end{align}
This general architecture allows to specifically choose the trade-off between giving more importance to the shared or the cluster specific part of the function. 
For the remainder of the paper, we consider the setting where $g_{\Psi_j}$ only consists of one fully connected layer, and the rest of the encoder is shared. This choice results in the model with the smallest number of parameters, while still preserving the overall idea of tensorization. The underlying idea is that even between data from different clusters, the fundamental structures would be shared and learned by the shared encoder. The combination of those features is then learned specific to the cluster.

\textbf{Optimization.}
Before going into the specific algorithms, we outline the details for the optimization of $\mathbf{PT}$ in \eqref{eq: PT General}.
The parameters of the model can be split into two main aspects: firstly, the parameters of the embedding functions (e.g., the weights of the neural networks) and secondly, the cluster assignment matrix $S$ which is unique to the cluster specific representation learning setup. For this part, we follow \cite{esser2023improved} and start by \emph{initializing} the parameters $\{ \Psi\}^k_{1},\Omega$ randomly and $S$ according to $k$-means ++ \cite{David2007kmeans}.
Subsequently, we iteratively repeat the following two steps until convergence:
\begin{enumerate}
    \item \emph{Update the parameters} $\{ \Psi\}^k_{1},\Omega$, for the combined and shared encoder using a GD step under fixed $S$.
    \item \emph{Update the cluster assignment} $S$ using a Lloyd’s step\footnote{Here the Lyod's step solves the linear problem on $S$ assuming all other parameters are fixed.} on a strict cluster assignment, that is, a data point $i$ can only belong to one cluster $j$ and thus $S_{j,i} \in \{0,1\}$.
\end{enumerate}


\textbf{Inference.}
In practice, we can optimize \eqref{eq: PT General} to obtain a cluster specific embedding for a given dataset. However, in some cases, we are also interested in the embedding of a new data point $x^*$. 
%
To do so, we first need to decide on the cluster assignment $s$ for the new data point $x^*$, specifying the embedding function $g_{\Psi_j}$ to be considered for $x^*$.
In order to find the cluster assignment $s$, it is natural to assign $x^*$ to the cluster that gives the smallest loss.
%
Following this idea, we define $s$ given the trained parameters $\{\Psi\}^k_{1},\Omega$ as: 
\begin{align}\label{eq:new datapoint}
    {s} =& \arg\min_j  \mathcal{L}\left(g_{\Psi_j}\left(g_\Omega(x^*)\right)\right)
\end{align}
and then compute the embedding as $g_{\Psi_{{s}}}\left(g_\Omega(x^*)\right)$.
This general approach can be applied to all considered optimization problems to obtain the embedding for a new data point.

\textbf{Evaluation metrics.} For evaluating clustering, we use the Adjusted Rand Index (ARI) values in $[-1, 1]$ \cite{hubert1985ARI} between the true labels and the predictions. Clustering with ARI $1$ implies predicted clusters are the same as the true partition, and $0$ implies the predicted clustering is random.
We report the Mean Squared Error (MSE) for assessing de-noising between the clean and noisy data (lower is better).
For evaluating the embeddings, there is no established metric to use directly and we discuss the relevant intuition in the corresponding sections.



\section{Cluster Specific Autoencoders}
\begin{figure*}[ht]
    \centering
    \includegraphics[width=0.32\linewidth]{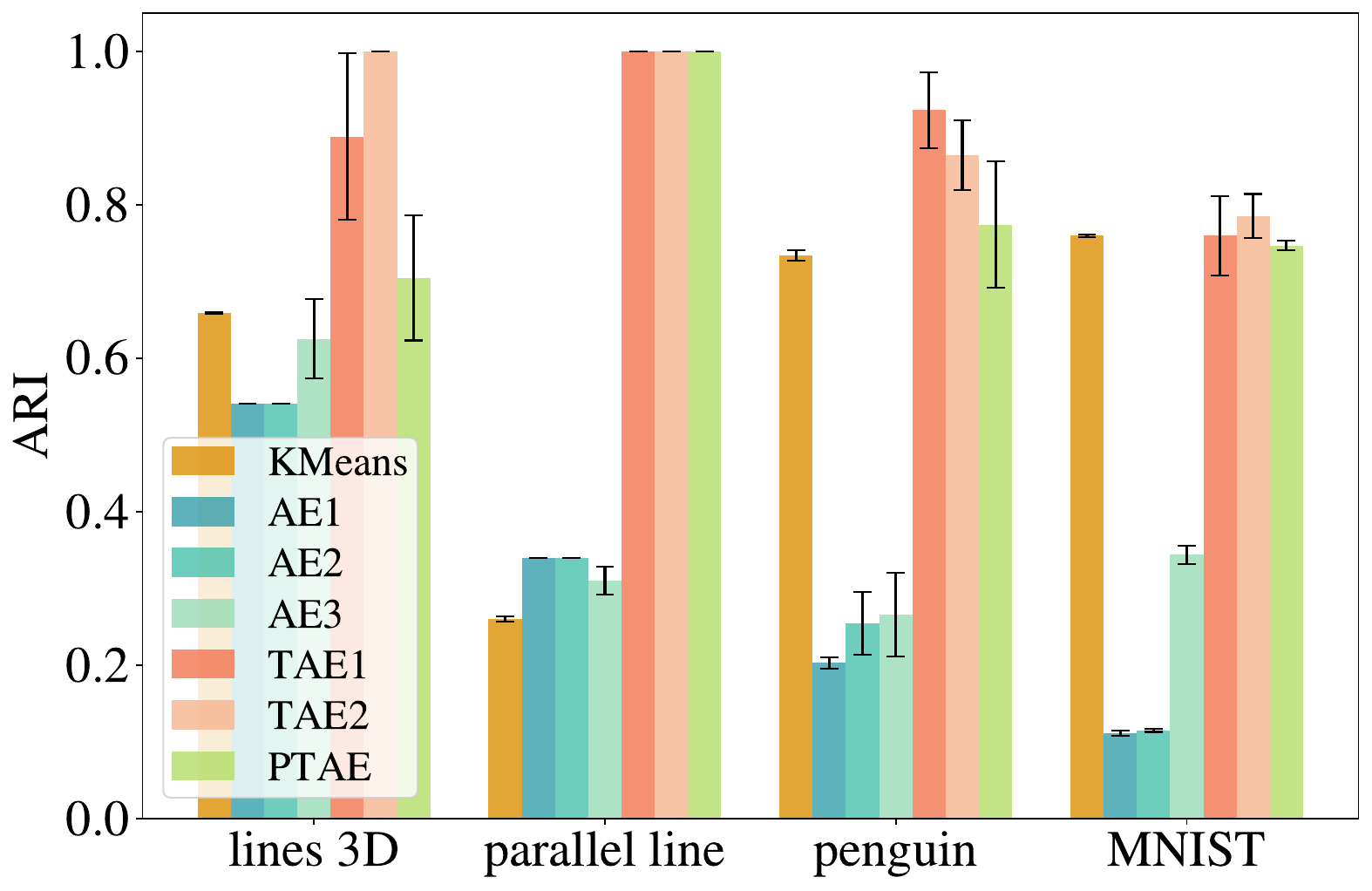}
    \includegraphics[width=0.32\linewidth]{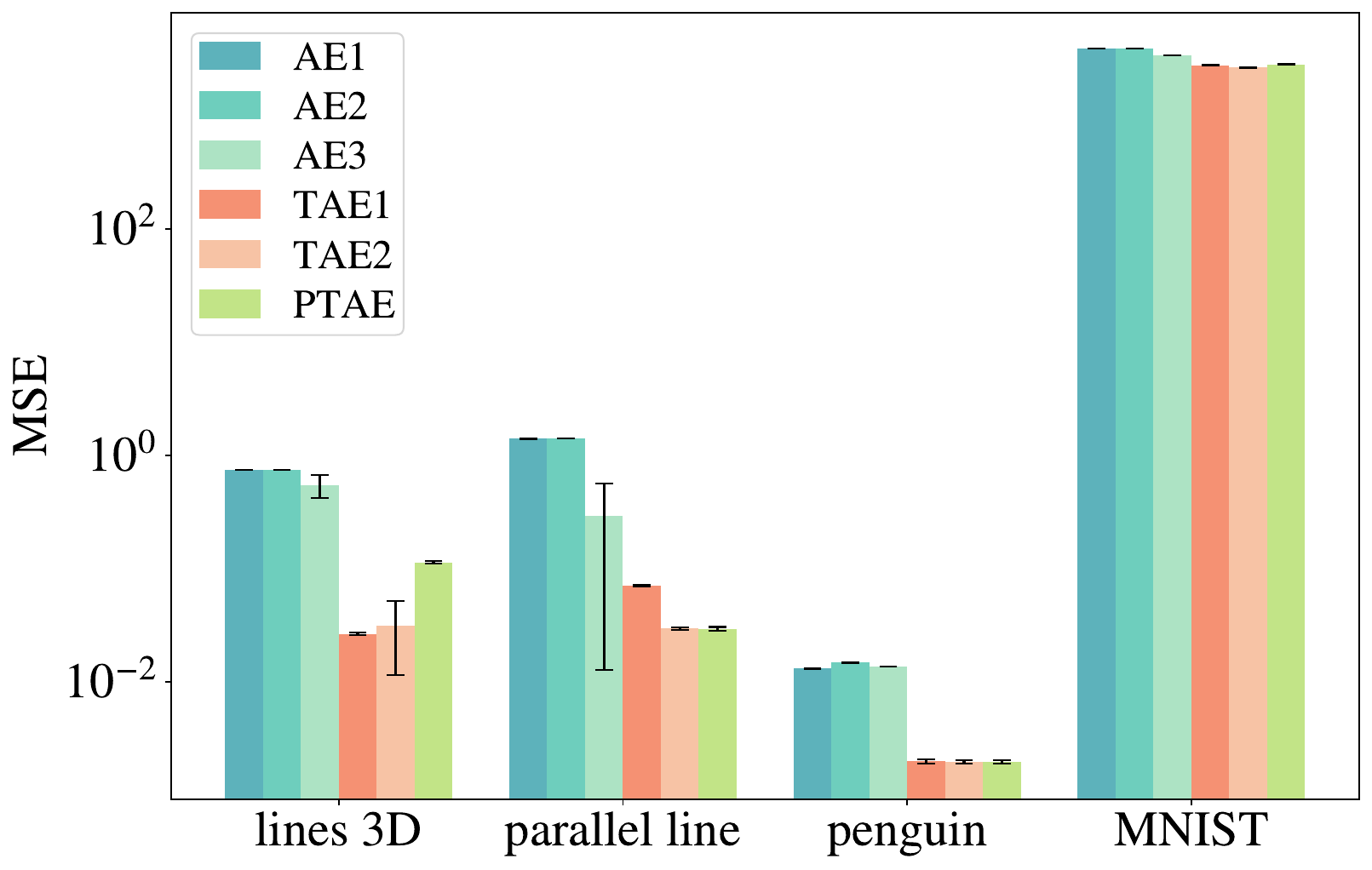}
    \includegraphics[width=0.32\linewidth]{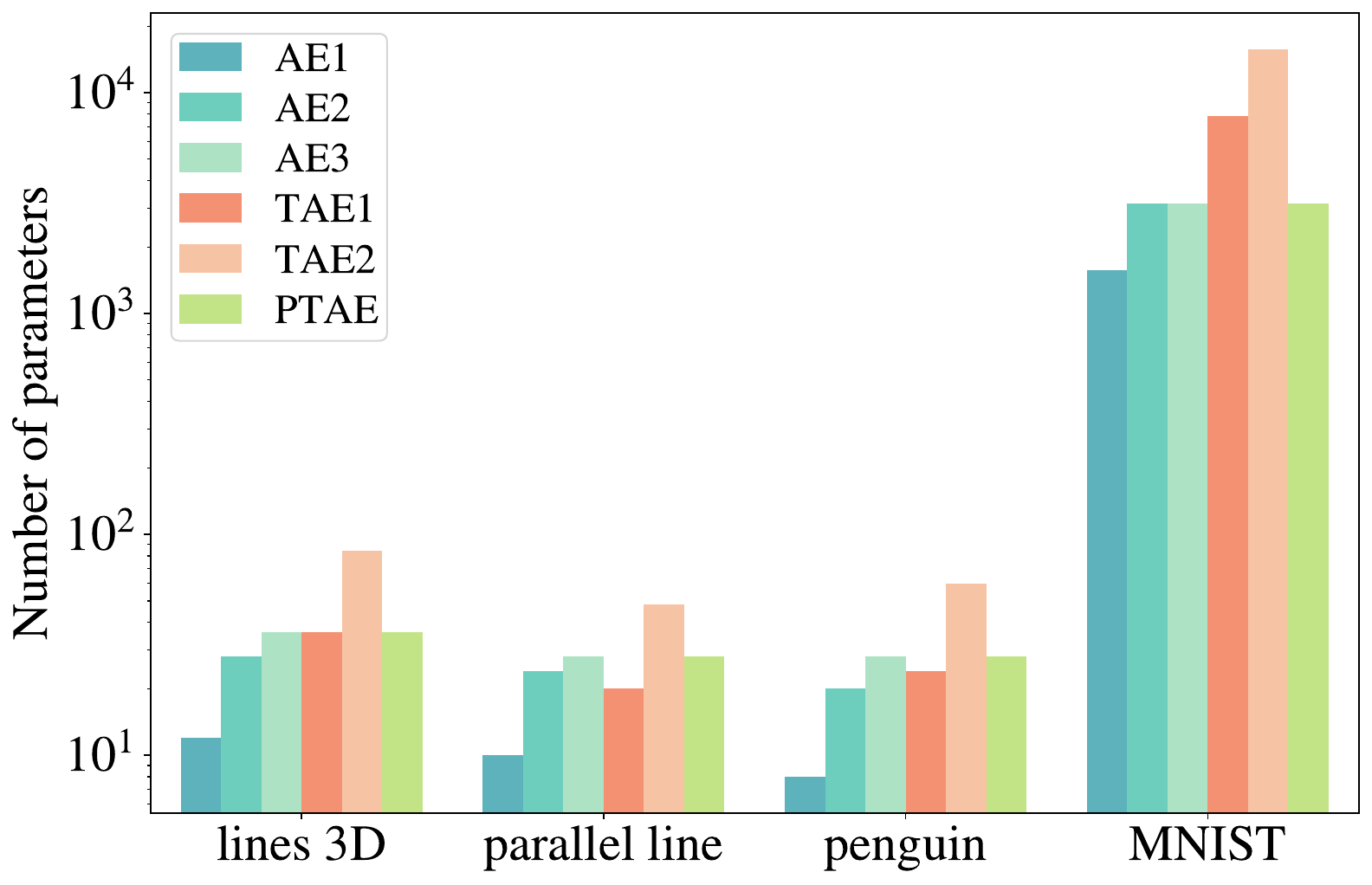}
    \caption{
    Performance of AE, TAE and PTAE.
    \textbf{Left:} clustering obtained by different models with k-means as benchmark. Plotted is ARI (higher is better). 
    \textbf{Center:} de-noising of different models. MSE in log-scale (lower is better). 
    \textbf{Right:} number of parameters in each considered model.}
    \label{fig:TAE}
\end{figure*}

We now consider the general setting of learning cluster specific representations outlined in the previous section and apply it to a range of commonly used representation learning algorithms.

    One of the most common approaches in deep learning for obtaining latent representations is Autoencoders\footnote{For a comprehensive introduction see \cite{hinton2006reducing}.} \cite{Kramer1991AIChE_AE} where the idea is to map the data into a lower dimensional space using an \emph{encoder} and then back into the original space using a \emph{decoder}. In most cases, both the encoder and decoder are defined by neural networks, which are optimized by minimizing the \emph{reconstruction error} through gradient-based methods.
 They have shown great success in a wide range of applications such as image de-noising \cite{buades2005review}, clustering \cite{yang2017towards} and natural language processing \cite{zhang2022survey}.

\subsection{Model Definition}
We consider the cluster specific model as defined in \eqref{eq: PT General} and extend it to the reconstruction based loss setting in this section. The partial tensorization is applied to the \emph{encoder and  decoder} such that the new overall model for the \emph{Partial Tensorized Autoencoder (PTAE)} is defined as
\begin{align}\label{eq:PTAE}
\mathbf{PTAE:}&\min_{\substack{\{\Phi_j, \Psi_j\}^k_{j},\\\Theta,\Omega,S}}
\sum^n_{i=1}  \sum^k_{j = 1}S_{j,i}\left[\norm{\tilde{x}_{i,j} - \hat{x}_{i,j}
}^2 
- \lambda p(\tilde{x}_{i,j})
\right]\nonumber\\
&\text{s.t. ~ } 1_k^TS = 1_n^T,\ S_{j,i} \geq 0,\\
&\text{and ~ }\hat{x}_{i,j}:=f_{\Theta}(f_{\Phi_j}(g_{\Psi_j}(g_{\Omega}(\tilde{x}_{i,j}))))\nonumber
\end{align}
with $\tilde{x}_{i,j}:=x_i-c_j$ where $c_j$ is the cluster center of $j$ computed using the data belonging to cluster $j$.
The PTAE architecture is illustrated in Figure 1 in Appendix.
As in \eqref{eq: PT General}, $S$ is a $k \times n$ matrix, such that $S_{j,i}$ is the probability that data point $i$ belongs to class $j$. Finally $g_{\Psi_j}( \cdot )$ and $f_{\Phi_j}( \cdot )$ are the cluster $j$ specific \emph{encoder} and \emph{decoder} functions, respectively. 
In addition to the standard AE formulation, we consider a regularized version by regularizing the embedding using $\lambda p(\tilde{x}_{i,j})$ where $\lambda$ determines the strength of the penalty.
Specifically, we consider $p(\tilde{x}_{i,j}):=\norm{g_{\Psi_j}(g_{\Omega}(\tilde{x}_{i,j}))}^2$ to be a $k$-means penalty on the embedding to enforce a cluster friendly structure in the latent space (similar to the one proposed in \cite{Bo2017k_means_friendly}).
Our PTAE definition is similar to the fully tensorized version presented in \cite{esser2023improved}. The fully tensorized AE (TAE) is obtained by defining the model only by shared encoders/decoders, such that $\hat{x}_{i,j}:=f_{\Phi_j}(g_{\Psi_j}(\tilde{x}_{i,j}))$.

\subsection{Numerical Evaluation}
We now empirically analyze the performance of the above defined $\mathbf{PTAE}$. As a reference, we also consider the simpler \emph{Tensorized Autoencoder (TAE)} as defined in \cite{esser2023improved}. 
We evaluate the models on clustering and de-noising tasks and observe that PTAE outperforms standard AE by a considerable margin and gives comparable performance to TAE with much reduced model complexity.

\textbf{Encoder \& decoder definition.} We define the following models by specifying the encoder and decoder in \eqref{eq:PTAE}. 
As a reference, we define the following standard AEs:
\emph{AE1:} one layer AE with embedding dimension $h=1$;
\emph{AE2:} two layer AE with hidden layer dimension $2$ and embedding dimmension $h=1$;
\emph{AE3:} two layer AE with hidden layer dimension $2$ and embedding dimmension $h=1 \times C$ where $C$ is the number of true clusters in the dataset;
We define cluster specific fully tensorized version TAE with $k=C$, following \cite{esser2023improved}:
\emph{TAE1:} one layer AE with embedding dimension $h=1$;
\emph{TAE2:} two layer AE with hidden layer dimension $2$ and embedding dimmension $h=1$;
Finally, we define a partial tensorized version \emph{PTAE} with similar architecture as \emph{TAE2} with only tensorization of the embedding layer.
Note that \emph{AE3} and \emph{PTAE} has the same model complexity (number of parameters).

\textbf{Clustering.} We compare the obtained clustering by i) k-means on the original data ii) k-means on the embedding obtained by standard AE and iii) the cluster assignment obtained by TAE and PTAE through $S$. We illustrate this comparison in Figure~\ref{fig:TAE} (left) for several datasets with $C$ number of true clusters\footnote{Datasets: \emph{parallel lines} $(d=5,n=150,C=2)$ \cite{esser2023improved}, axis parallel \emph{lines in 3D} $(d=6,n=300,C=3)$ \cite{esser2023improved}, \emph{MNIST} $(d=784,n=1000,C=5)$ sampled $200$ samples from classes $\{0,1,2,3,4\}$ \cite{mnist} and \emph{penguin} $(d=4,n=334,C=3)$ \cite{Penguins}. Note that \emph{parallel lines} and \emph{penguin} are toy and real dataset, respectively, illustrating the simpson's paradox.}.
Firstly, we observe that all three standard AE models perform similarly, and the increase in model complexity does not reflect in the performance of the model. It is interesting to note the baseline k-means is outperforming AEs in most cases. Comparing the standard AEs and the tensorized versions, we observe that the TAEs outperform the AEs consistently, demonstrating the significance of tensorization albeit increase in the model complexity. On the other hand, PTAE performs comparable to the TAEs with much lesser model complexity and outperforms AEs with almost similar model complexity, thus illustrating the power of partial tensorization.

\textbf{De-Noising.} For the same models and datasets considered for clustering, we evaluate the de-noising performance, measured using the MSE of the reconstruction. Here the noisy version of a sample $x_i$ is generated as $x_i^{\text{noise}}:=x_i+\varepsilon,~\varepsilon\sim\mathcal{N}(0,0.1)$.
Figure~\ref{fig:TAE} (middle) shows that TAEs significantly reduce the MSE compared to standard  AEs. Note that the results are presented on a log-scale. In addition we again observe that the PTAE performs on par with the TAEs, further demonstrating the capabilities of partial tensorization.

\textbf{Model complexity.} Finally, we compare the number of parameters for the considered AEs in Figure~\ref{fig:TAE} (right). It clearly shows that \emph{PTAE does not add more complexity compared to the standard AE, while achieving significantly better performance in clustering and de-noising.}
For smaller models (e.g. the ones used for \emph{lines 3D}, \emph{parallel lines} and \emph{penguin}) the difference is less significant, but more pronounced for the network considered for \emph{MNIST}. 

\textbf{Additional results in appendix.} We provide experiments for additional datasets as well as a runtime analysis.

\section{Cluster Specific Variational Autoencoders}

Starting from AE we can extend our analysis to learning a probabilistic representation of the data. This has been established through Variational Autoencoders\footnote{For a comprehensive introduction see for example \cite{doersch2016tutorial,kingma2019introduction}.} (VAE) \cite{kingma2013auto}, a generative model that employs neural networks to learn the underlying distribution of data. It combines the principles of variational inference and autoencoders and has numerous application in image generation such as face generation, style transfer, and super-resolution \cite{kingma2013auto}, text generation \cite{bowman2015generating}, anomaly detection
\cite{an2015variational} and representation Learning \cite{higgins2017beta}.
\subsection{Model Definition}
Formally, let $ {x} $ be the observed data and $ {z_j} $ the latent variables corresponding to cluster $j$. The VAE consists of two main components:
the \emph{encoder} is given by $ q_{\Psi_j,\Omega}\left({z_j}|{x_i}\right) $, a neural network parameterized by $\Psi_j,\Omega $, which approximates the posterior distribution of the latent variables given the observed data.
Similarly the \emph{decoder} is defined by $ p_{\Phi_j,\Theta}\left({x_i}|{z_j}\right) $, a neural network parameterized by $\Phi_j,\Theta$, which models the likelihood of the observed data given the latent variables.
The TVAE optimizes the Evidence Lower Bound (ELBO) on the marginal likelihood:
\begin{align}\label{eq: VAE objective}
\mathbf{TVAE}:&\min_{{\{\Phi, \Psi\}^k_{1},\Theta,\Omega,S}}
\sum_{i=1}^n\sum_{j=1}^k S_{i,j}
\left[  \text{Rec}_{i,j} - \text{Reg}_{i,j}\right]\\
&  \text{s.t. ~ } 1_k^TS = 1_n^T,\ S_{j,i} \geq 0,\nonumber\\
\text{Rec}_{i,j}:& ~ \mathbb{E}_{q_{\Psi_j,\Omega}({z_j}|{x_i})}\left[\log p_{\Phi_j,\Theta}\left({x_i}|{z_j}\right)\right]\label{eq:vae_rec}\\
\text{Reg}_{i,j}:& ~ \text{KL}\left(q_{\Psi_j,\Omega}\left({z_j}|{x_i}) \middle\| p_{\Phi_j,\Theta}({z_j}\right)\right) \label{eq:vae_reg}
\end{align}
where $ \text{KL}(\cdot\|\cdot)$ denotes the Kullback-Leibler divergence, and $ p({z_j}) $ is the prior distribution over the latent variables, typically a standard Gaussian $ \mathcal{N}({0}, \mathbb{I}) $.
In addition, \eqref{eq:vae_rec} is referred to as the \emph{Reconstruction Term} and \eqref{eq:vae_reg} as the \emph{Regularization Term}. 

\textbf{Encoder \& decoder definition.} 
Note that the latent space is now a probabilistic space, more specifically a Gaussian, and we are now interested in learning those parameters using neural networks.
Again let $\tilde{x}_{i,j}:=x_i-c_j$ where $c_j$ is the cluster center of $j$ computed using the data belonging to cluster $j$.
Firstly, we define a embedding layer $g_\Omega:\mathbb{R}^d\rightarrow\mathbb{R}^{l}$ that is shared for the mean and variance of the latent space and then from there  and using this the TVAE
learns the \emph{cluster specific} mean and variance of the embedding:
where $\mu_{i,j} = g_{\Psi_j^\mu}(g_\Omega(\tilde{x}_{i,j}))):\mathbb{R}^d\rightarrow\mathbb{R}^{h}$ learns the mean and $\log(\sigma_{i,j}^2) = g_{\Psi_j^\sigma}(g_\Omega(\tilde{x}_{i,j}))):\mathbb{R}^d\rightarrow\mathbb{R}^h$ the variance of the latent distribution.
%
The decoder maps the latent variable $z$ back to data space using a cluster specific $f_{\Phi_j}:\mathbb{R}^h\rightarrow\mathbb{R}^l$ and shared $f_{\Theta}:\mathbb{R}^l\rightarrow\mathbb{R}^d$ decoder: $\hat{\mu}_{i,j} = f_{\Theta}(f_{\Phi_j}({z}_{i,j}))$.
Since the latent space is probabilistic, we do not have direct access to $z$. However, we can use the following sampling procedure to obtain $z$.

\textbf{Sampling.} To sample the latent distribution we can take advantage of the \emph{reparameterization trick},
$
    z_{i,j} = \mu_{i,j} + \sigma_{i,j}^2 \odot \varepsilon,
$
where $\varepsilon\sim\mathcal{N}(0,\mathbb{I})$ and $\odot$ is Hadamard product.

\textbf{Reconstruction term and optimization.} In practice the expectation in \eqref{eq: VAE objective} can be approximated by the empirical mean using a Binary Cross Entropy loss\footnote{Formally we can define BCE in the tesorized context as:
$
-\frac{1}{n} \sum_j^k S_{i,j}\sum_{i=1}^n[\tilde{x}_{i,j} \log (\phi(\hat{\mu}_{i,j}))+(1-\tilde{x}_{i,j}) \log (1-\phi(\hat{\mu}_{i,j}))]
$
where $\phi(\cdot)$ is the sigmoid function.
This formulation of the reconstruction is not a unique choice. An alternative formulation is using MSE instead of BCE. We present results for this in the supplementary material.}.
The parameters of the model are optimized using GD on the reconstruction and regularization term.







\subsection{Numerical Evaluation}
\begin{figure}[t]
    \centering
    \includegraphics[width=0.23\linewidth]{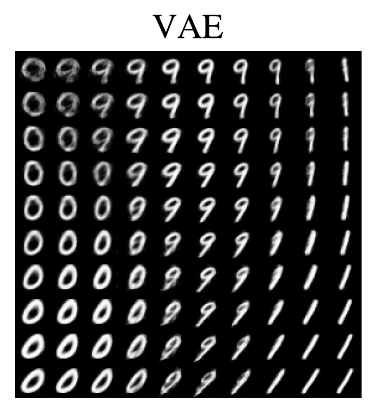}
    \includegraphics[width=0.23\linewidth]{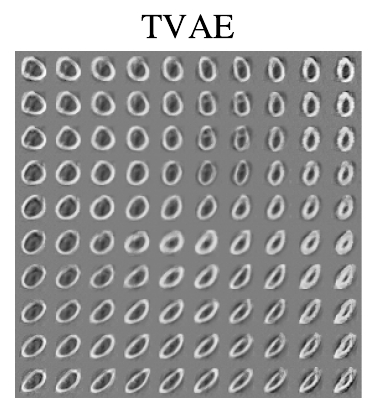}
    \includegraphics[width=0.23\linewidth]{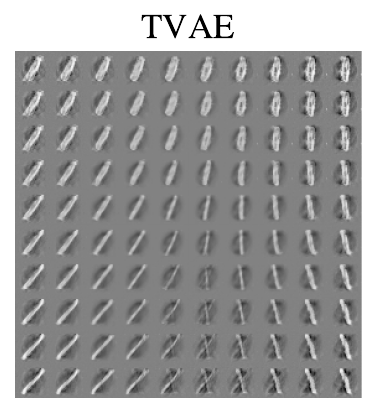}
    \includegraphics[width=0.23\linewidth]{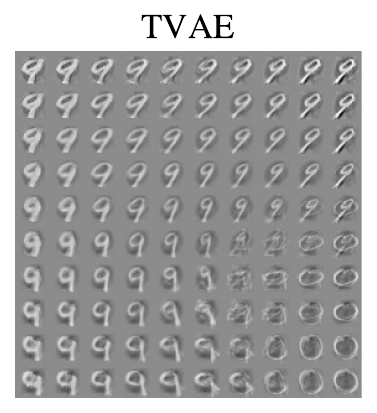} \\
    \hspace{-0.5em}
    \includegraphics[width=0.23\linewidth]{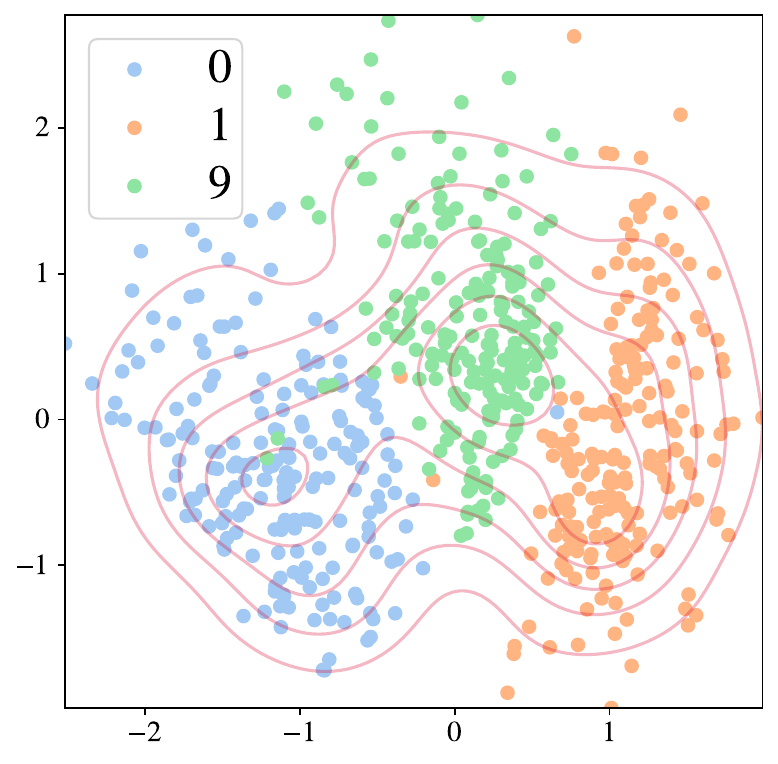}
    \hspace{-0.1em}
    \includegraphics[width=0.23\linewidth]{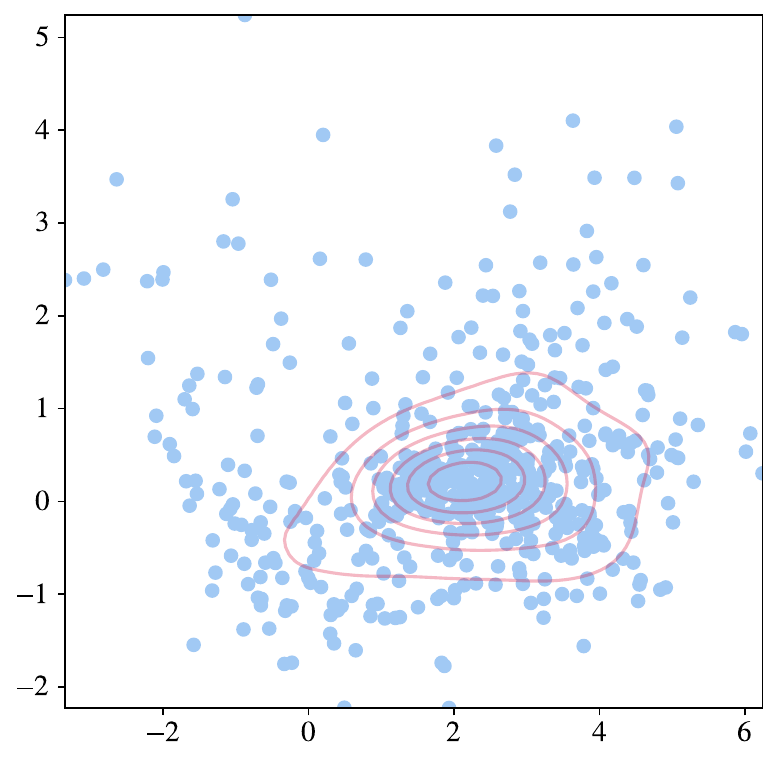}
    \hspace{-0.1em}
    \includegraphics[width=0.23\linewidth]{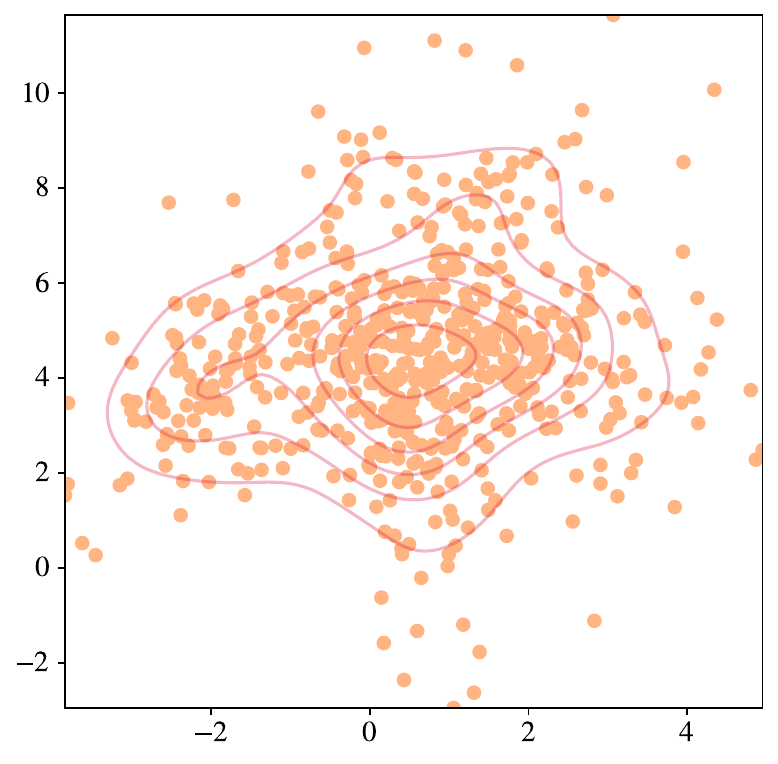}
    \includegraphics[width=0.23\linewidth]{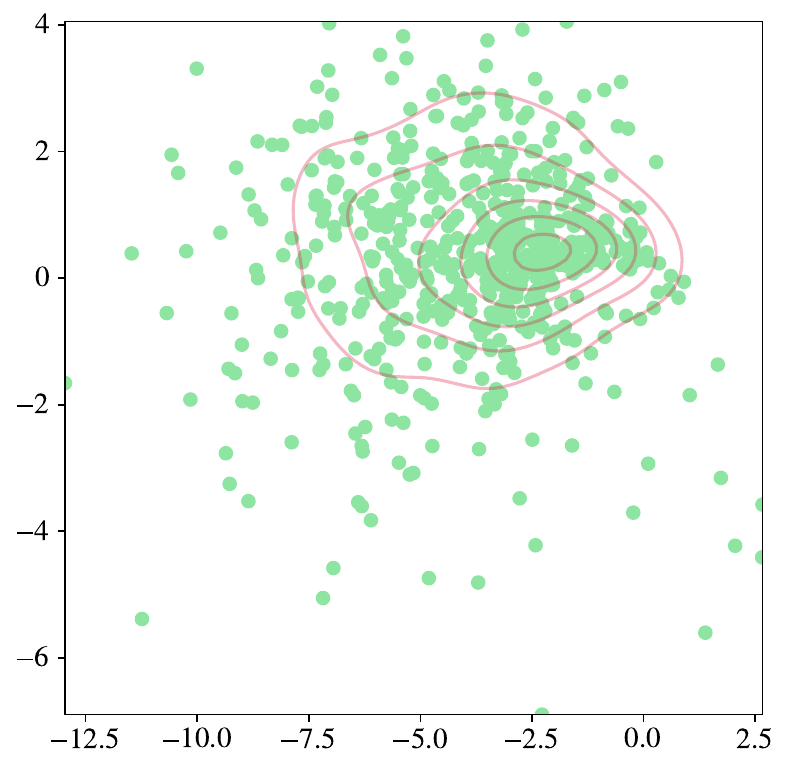}\\
    \hspace{-0.5em}
    \includegraphics[width=0.23\linewidth]{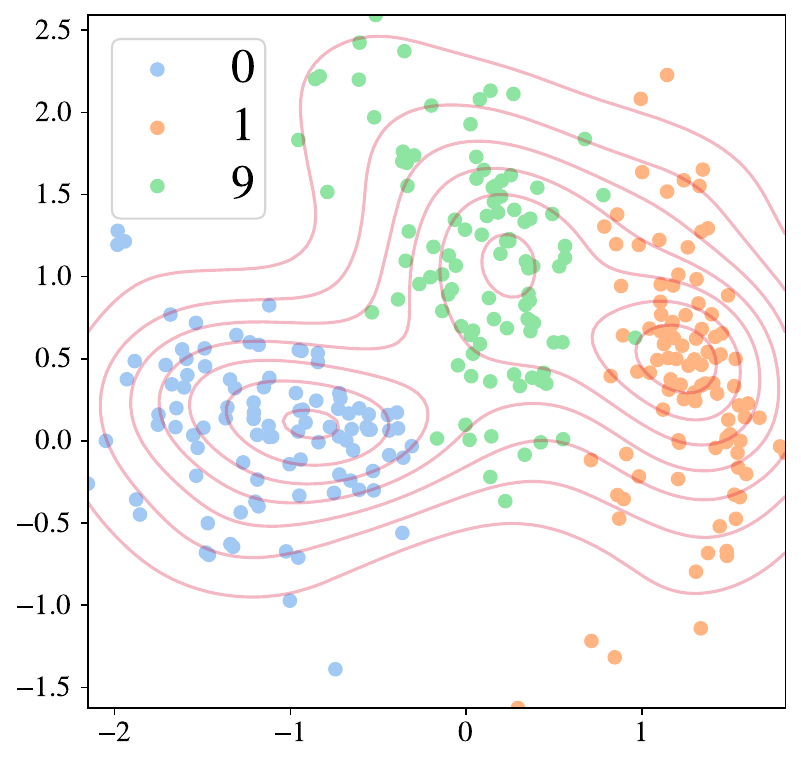}
    \hspace{-0.1em}
    \includegraphics[width=0.23\linewidth]{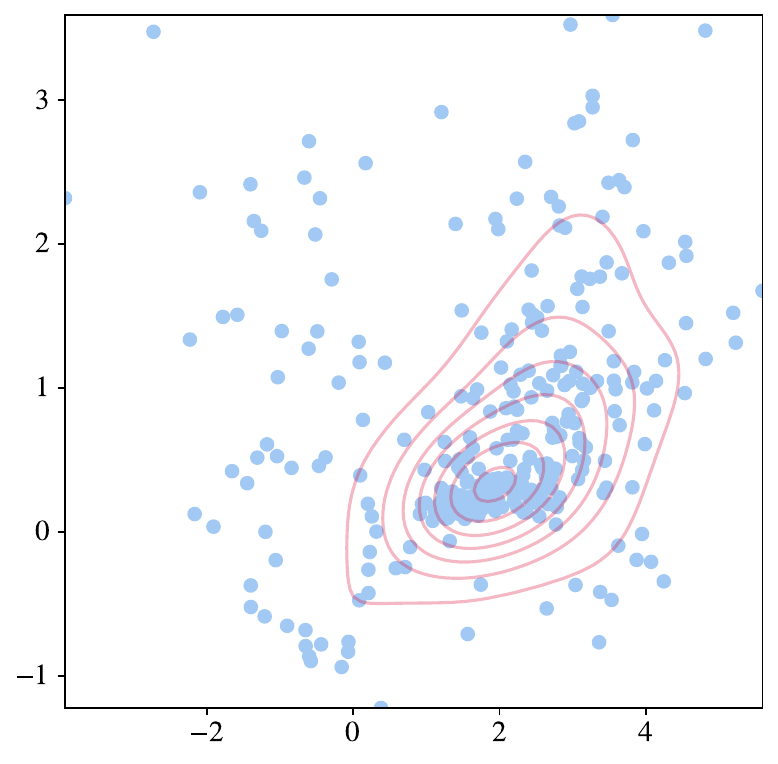}
    \hspace{-0.1em}
    \includegraphics[width=0.23\linewidth]{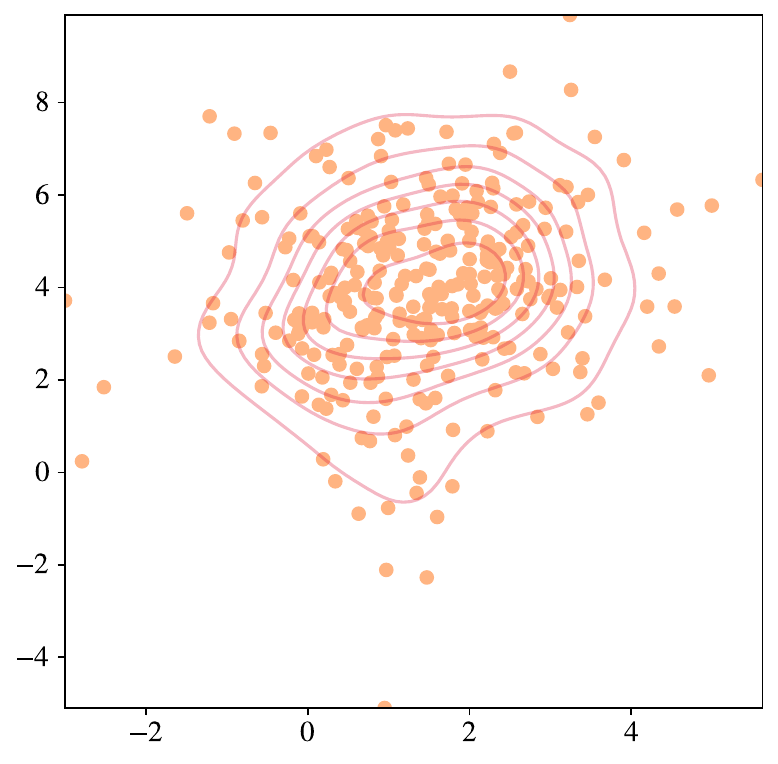}
    \includegraphics[width=0.23\linewidth]{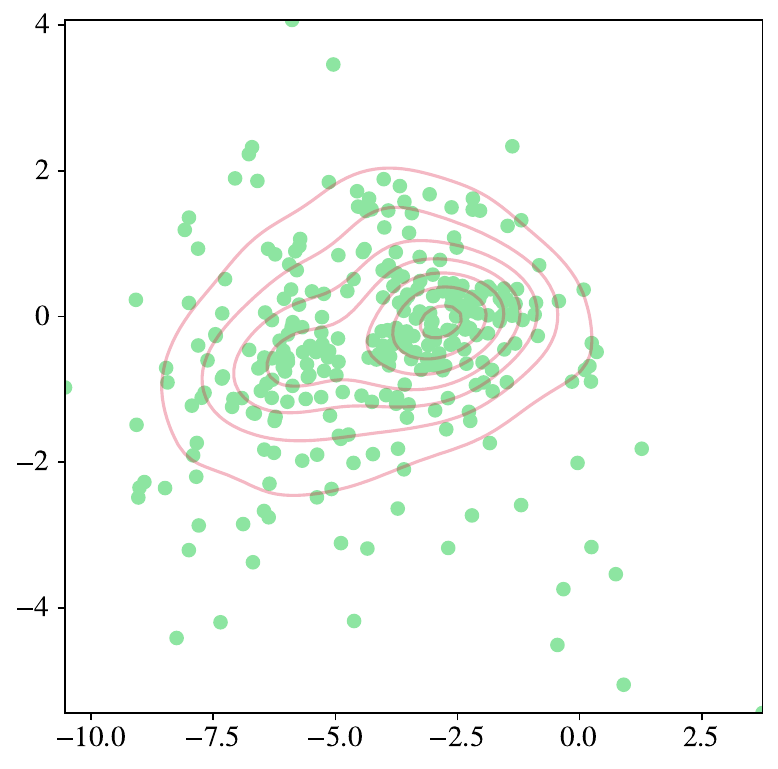}
    \caption{
    \textbf{Top row: samples from the latent space.} The very left plots shows samples from the latent space of the standard VAE. The following three plots show samples from the cluster specific latent spaces of the TVAE.
    \textbf{Middle row: embedding of training samples.}
    Embedding of training data-points using the standard VAE and TVAE. Plotted is the obtained mean as well as the contour lines of Gaussian model.
     \textbf{Top row: embedding of test samples.} Same setting as the middle row but for unseen data-points.
    }
    \label{fig:vae_latent_space}
\end{figure}
We define a simple tensorized VAE model by setting $g_{\Psi_j^\mu},g_{\Psi_j^\sigma}, g_\Omega,f_{\Theta}$ and $f_{\Phi_j}$ each to be one layer where the hidden dimension is $200$. Furthermore, for illustrative purposes, let the embedding dimension be two. 

\textbf{Sampling the latent space.} 
One especially interesting property of VAEs for our analysis is that the probabilistic nature of the latent space allows for a direct visualization of the learned representations through sampling of the latent distribution.
This is illustrated in Figure~\ref{fig:vae_latent_space} (top row), where we show the learned latent space for the digits $0,1$ and $9$ from the MNIST dataset. On the left, we plot the latent space for the standard VAE, where there are sections for the different digits and interpolations between them. 
The following three plots show the three latent spaces for the TVAE. It is clear to see that if we sample from the different latent spaces now, each of them only contains digits from one of the three classes. We can also observe that when considering separate embeddings, the expressiveness of the considered styles increases (for \emph{one} and \emph{nine}, the embedding of tilts in both directions is now learned in comparison to the standard VAE). In addition, we note that the standard VAE results in a higher contrast in the sampled images. However this can be explained by the fact that we center the data and could be removed by re-scaling the outputs, which we do not perform at this point to preserve a fair comparison of the results.

\textbf{Embedding of data-points.}
In a similar fashion to the above analysis, we analyze the embeddings obtained by the training data and new data-points. To do so, we again consider the setting of embedding $200$ digits each from classes $0,1$ and $9$ from the MNIST dataset.
To obtain the embedding, we use \eqref{eq:new datapoint} to assign the data-point, $x_i^*$, to the appropriate encoder.
Figure~\ref{fig:vae_latent_space} second row illustrates the mean, $\mu_{i,j}$, obtained for the training data-point and similarly, Figure~\ref{fig:vae_latent_space} second row for unseen data-points. We observe that for the standard VAE, all embeddings are grouped by class but overall are constrained to a standard Gaussian (due to the regularization term in the objective function). On the other hand for the TVAE each class learns an embedding, following a Gaussian distribution, however due to the different latent spaces the TVAE allows to learn different mean and variances for each of the classes. This allows to better capture the structure of the data.


\textbf{Additional results in the appendix.} We provide experiments for additional datasets including Fashion MNIST, different reconstruction objectives and a setting where the true number of classes and the number of embedding functions differs. 

\section{Cluster Specific Contrastive Loss}
In the previous two sections we considered models that are based on mapping the data into lower dimensional (probabilistic) spaces and then optimize the encoding and decoding functions through a reconstruction loss. We now show that the overall idea of tensorization can be extended beyond this setting to losses that are defined directly on the embedding.
More specifically we consider the \emph{Self Supervised learning (SSL) \cite{bromley1993signature} setting}
which builds on the idea that data-points which are ``similar'' in the features space should be mapped close to each other in the latent space. This main idea is formalized through \emph{Contrastive Losses (CL)}\footnote{For detailed introductions to contrastive learning refer to \cite{ericsson2022self,liu2021self}.}.
The inter-sample relations are constructed through data-augmentations of $X$ known to preserve input semantics such as additive noise or random flip for an image \cite{kanazawa2016warpnet}. This idea has been highly successful in domains such as computer vision \cite{chen2020simple,Jing2019SelfSupervisedVF} and natural language processing \cite{ misra2020self,BERT2019}.

\subsection{Model Definition}
\textbf{Data setting.} In contrastive learning, the idea of mapping semantically similar pairs close to each other is implemented through the use of positive and negative samples. Consider a reference sample $x$, the goal is to learn an embedding that results in embedding of $x$ very similar to the embedding of positive sample $x^+$, and dissimilar to the embedding of negative sample $x^-$.
The positive sample is usually generated using semantic information of the data. In the context of images, this is achieved, for example, by adding noise or a horizontal flip for an image \cite{kanazawa2016warpnet, novotny2018self, gidaris2018unsupervised}.
The negative sample is chosen to be another independent sample from the dataset.



\textbf{Loss function.} To formalize the idea of mapping similar points close to each other in the latent space, we define the cluster-specific version of the contrastive loss \cite{Arora2019ATA}. For notational convenience, we suppress the parameters of the embeddings.






Given a set of anchor samples $\{x\}_{1}^n$, with corresponding positive samples $\{x^+\}_{1}^n$ and negative samples $\{x^-\}_{1}^n$, the model employs $k$ cluster-specific embedding functions, which can be combined into the overall \emph{Tensorized Contrastive Loss} objective:
\begin{align*}
\mathbf{TCL:} \min_{\{\Psi\}_{1}^k,\Omega,S}\frac{1}{n} \sum_{i=1}^{n} \sum_{j=1}^{k} S_{i,j} \left[ \left\langle z_{i,j}, z_{i,j}^- \right\rangle-\left\langle z_{i,j}, z_{i,j}^+ \right\rangle \right]\\
\text{s.t. ~ } 1_k^TS = 1_n^T,\ S_{j,i} \geq 0\text{ ~ with ~ }z_{i,j}:=g_{\Psi_j}\left(g_\Omega(x_i)\right).
\end{align*}
We consider $g_\Omega$ to be two convolutional layers followed by $g_{\Psi_j}$ being one fully connected, cluster specific layer. 
The architecture is illustrated in Figure 1 in Appendix.

\subsection{Numerical Evaluation}
\begin{figure}
    \centering
    \includegraphics[width=1\linewidth]{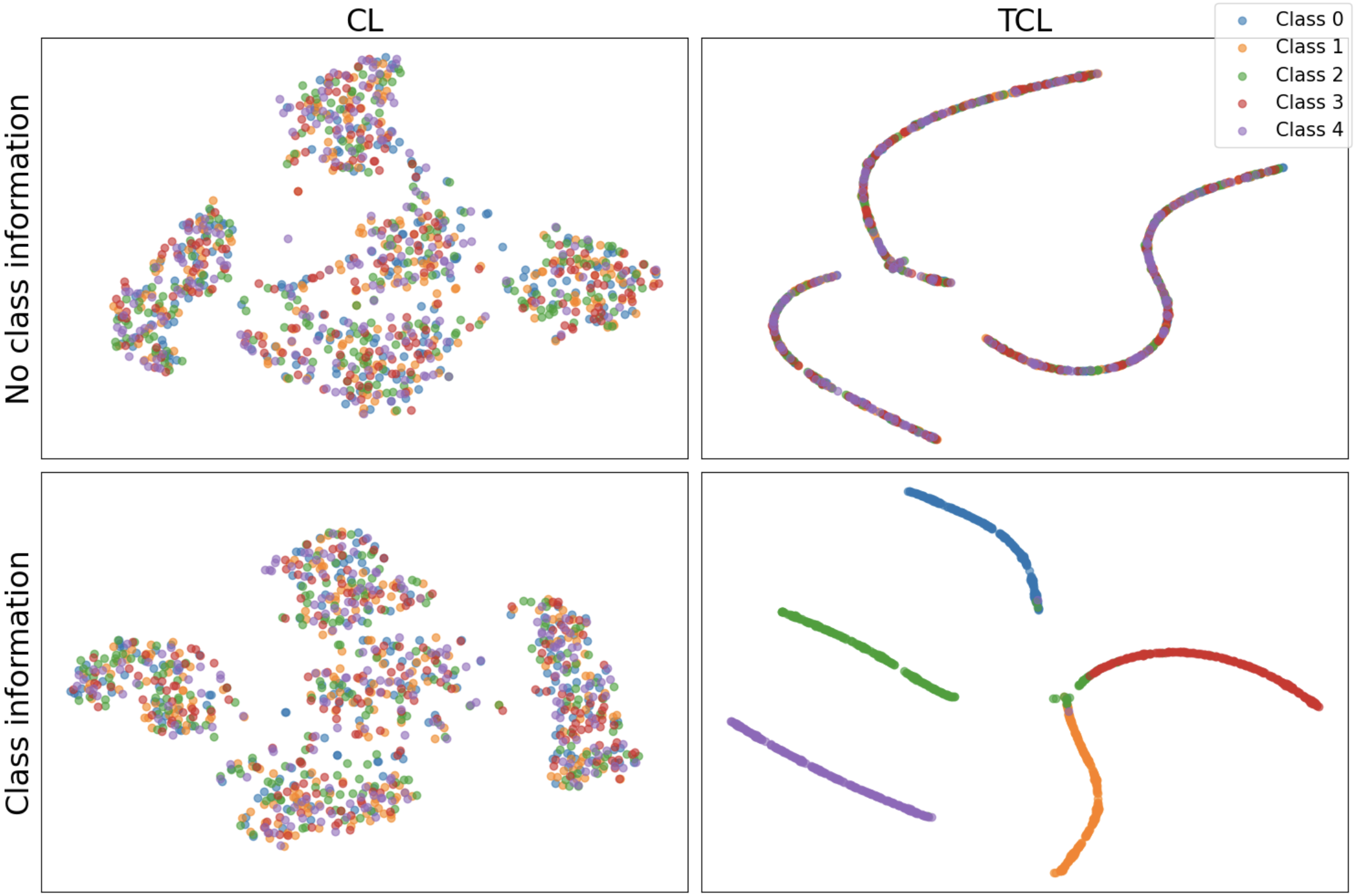}
    \caption{Illustration of the embedding obtained from the (tensorized) contrastive loss.
    \textbf{Columns:} \emph{left} shows standard CL setting and \emph{right} shows the TCL setup.
    \textbf{Rows:} \emph{top row} shows the embedding obtained without any class information and \emph{bottom row} shows the embedding obtained under class information.
    }
    \label{fig:SSL}
\end{figure} 


\textbf{Positive and negative sample generation.} As outlined previously an important difference between the unsupervised settings of the previous two sections and the SSL setup is the consideration of semantic similarities through the postive and negative samples, and a central question is how to define them. For the experimental setup, we consider learning embeddings using both supervised and unsupervised methods:
    for each sample $ x $, positive and negative samples are generated depending on whether class knowledge is available.
     In this work, we consider two approaches to generate the positive and negative samples:
\begin{enumerate}
    \item \textbf{Unsupervised.} 
    This is the most commonly considered setting \cite{ericsson2022self} where we do not assume any access to class information.
    \emph{Positive samples}, $ x^+ $, are generated by applying an augmentation; in our case ``elastic transformation'' \cite{1227801} to the image sample $x$. This transformation introduces controlled, non-linear distortions that preserve the underlying semantic content of the image, ensuring that the positive sample remains similar to the anchor.
    \emph{Negative samples}, $ x^- $, are selected based on the current cluster assignments $S$. The negative sample is chosen from a different cluster, ensuring that it is dissimilar to the sample $x$. Note that this is not possible in the standard CL setup where the image is usually an independent sample from the whole dataset.
    \item \textbf{Supervised.}  
    In this setting we now assume that there is access to class information, that can be employed to obtain better embeddings \cite{robinson2021contrastive,khosla2020supervised,zhang2021supporting}.
    \emph{A positive sample} $x^+$ is selected from the same class as the image $x$ but is a different instance from the dataset.
    \emph{The negative sample} $x^-$ is selected from a different class than the image $x$, ensuring a clear semantic distinction.
\end{enumerate}

\textbf{Evaluation.}  
The evaluation of the learned embeddings is performed by visualizing the $128$-dimensional embeddings in a two-dimensional space using t-SNE (t-distributed Stochastic Neighbor Embedding) \cite{vandermaaten08a}. This technique is effective for assessing how well the model has learned to separate different classes or clusters in the latent space.

The results are presented in Figure~\ref{fig:SSL}. Firstly comparing CL and TCL (left and right column) we observe that the latent structures learned differ significantly. While CL seems to learn connected point clouds, TCL learns curves in the embedding space. 
In the case when no class information is provided, the obtained latent structures have cluster structures, but they do not seem to align with the true labels for both CL and TCL. This is a common observation \cite{perrot2020near,chapelle2005semi,chang2017deep}, and while this is a problem in the context of clustering, when considering other tasks such as de-noising this is not an issue (as we show in the TAE section). Intuitively for a given data-set there might be several natural ways to cluster the data and the one learned might not align with the ``true labels''.
%
When comparing the setting with and without class information, CL shows slightly better separation of the clusters, however does not seem to group similar data points together. In contrast when applying TCL together with class knowledge (lower right), we observe a latent representation with similar classes grouped close together.









\section{Cluster Specific  Boltzmann Machine}
As a final model we analyze a traditional, energy function based model for representation learning and show that such settings can also benefit through cluster specific setting.
 Restricted Boltzmann Machines (RBM)\footnote{For a comprehensive introduction see \cite{fischer2014training}.} \cite{smolensky1986information} are a type of stochastic neural network that can learn a probability distribution over its set of inputs. It consists of two layers: a ``visible layer'' representing observable data, and a ``hidden layer'' representing latent variables. The key characteristic of an RBM is the restriction that there are no intra-layer connections; that is, the visible units are not connected to each other, and the hidden units are not connected to each other, only to units in the other layer. 
While simple in comparison to modern neural network approaches, RBMs have applications in feature extraction and data compression \cite{hinton2006reducing}, image and signal de-nosing \cite{vincent2008extracting}, anomaly detection \cite{choi2019deep} and pre-training in the context of deep belief network \cite{hinton2006fast}.

 \subsection{Model Definition}

We now formalize the above idea of RBM and outline the tensorized versions.
   In the context of RBM, the inputs are referred to as \emph{visible units},  $ {x_i} \in \mathbb{R}^d $.
   Furthermore, for each class $j\in \{1,\ldots,k\}$ we define \emph{hidden units}, $ {z_j} \in \mathbb{R}^h $, \emph{weights}\footnote{Note that in this simple model $g_{\Psi_j} = W_j$ and we do not consider a partial but only full tensorization.} $ W_j \in \mathbb{R}^{d \times h} $ as well as \emph{biases} (visible bias vector $ {a} \in \mathbb{R}^d $ and hidden bias vector $ {b_j} \in \mathbb{R}^h $).

   Assume $x_i$ belongs to class $j$. Then the energy of a configuration $({x_i}, {z_j})$ is given by
   $
   \mathcal{E}_{{W_j},a_j,b_j}({x_i}, {z_j}) = -{x_i}^\top {a_j} - {z_j}^\top {b_j} - {x_i}^\top W_j {z_j}.
   $
   Intuitively observe that the dot products in this expression show the ``alignment'' between visible and hidden units as well as bias terms.
   From there we can define the joint probability distribution over the visible and hidden units is:
     $
         P_{W_j}({x_i}, {z_j}) = {A_j}^{-1} \exp\left(-\mathcal{E}_{{W_j},a_j,b_j}({x_i}, {z_j}) \right)
     $
     where $A_j$ is the partition function:
    $
     A_j = \sum_{{x_i}} \sum_{{z_j}} \exp(-\mathcal{E}_{{W_j},a_j,b_j}({x_i}, {z_j}) ).
     $
The marginal probability of a visible vector $ {x_i} $ is:
     $
          P_{W_j,h_j}({x_i}) = {A_j}^{-1} \sum_{{z_j}} \exp(-\mathcal{E}_{{W_j},a_j,b_j}({x_i}, {z_j}) ).
     $
Finally we note that the goal of training an RBM is to maximize the likelihood of the training data. Therefore combining the above, and defining $P_{i,j}:=\log \left( P_{W_j,h_j}({x_i}) \right)$ the log-likelihood over the dataset and over all classes is given by:
   \begin{align*}
     \mathbf{TRBM:}  &\min_{\{W,a,b\}_1^k,S} \frac{1}{n}\sum_j^k \sum_i^n S_{i,j}\left[\left( P_{i,j}- \log A_j\right)\right]\\
   &\text{s.t. ~ } 1_k^TS = 1_n^T,\ S_{j,i} \geq 0.
   \end{align*}
   We can optimize the above expression by again iterative updating $\{W,a,b\}_1^k$ using GD and $S$.

 \begin{figure}
    \centering
    \includegraphics[width=0.32\linewidth]{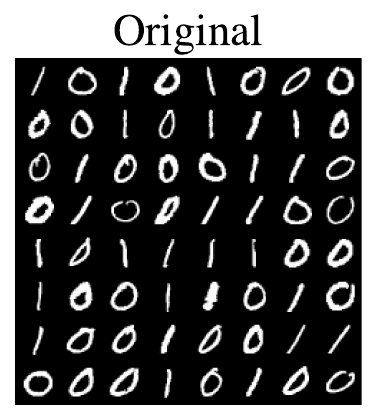}
    \includegraphics[width=0.32\linewidth]{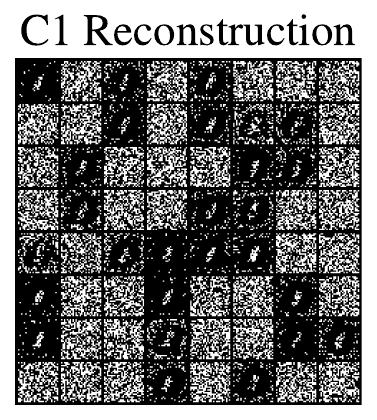}
    \includegraphics[width=0.32\linewidth]{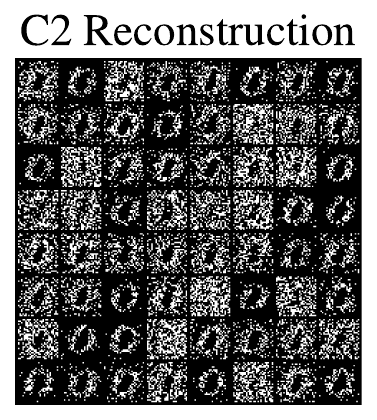}
    \caption{Reconstruction through TRBM.
    \textbf{Left:} true samples
    \textbf{Middle/Right:} reconstruction for class one and two.
    }
    \label{fig:rbm}
\end{figure}

\subsection{Numerical Evaluation}
We again consider the MNIST dataset and look at the cluster specific reconstructions obtained for digits from class \emph{zero} and \emph{one}. 
Figure~\ref{fig:rbm} (left) shows the input digits we are interested in reconstructing, and the subsequent two plots show the cluster specific reconstructions. We observe that the two embedding functions indeed \emph{only} learn the embedding specific to one class. Compare the middle or right with the left plot: we observe that for the specified class, the model manages to reconstruct the given images, while reconstructions of digits from the other class only result in random noise.


\section{Discussion}
In this final discussion, we address some conceptual questions on the idea of cluster specific representation learning.
We address the \emph{need} for tensorization by discussing the following questions:

\textbf{Can we use a larger embedding size to learn cluster specific structures instead of tensorization?} 
While this might be possible in specific instances, the example of Figure~\ref{fig:simpsons} shows a counter example. Even when increasing the latent space, the linear AE would still only learn the principal components of the full data and could not recover cluster specific representations.

\textbf{Can we use more complex models to learn cluster specific structures instead of tensorization?} 
We note that a sufficiently complex model would be able to interpolate between the cluster specific representations. However, it has been observed in several works that machine learning algorithms have a bias towards learning \emph{simple functions} \cite{domingos1999role,vapnik1998statistical}. Therefore, while it is possible to construct a model that \emph{can} learn a cluster specific representation, it is less likely that the model \emph{would actually learn it}.

Secondly, we discuss the fact that the approach is based on the number of clusters being known.

\textbf{Unknown number of clusters.} 
We now discuss heuristically the influence of wrongly estimating the number of clusters. 
\emph{Overestimate the number of clusters:} while this would introduce a computational overhead, the additional embedding functions would simply learn copies of one of the cluster specific embedding functions.
\emph{Underestimate the number of clusters:} in this case most similar clusters would be combined to one of the encoding functions.

Thus, cluster specific representation learning is a promising approach to capture the intrinsic structure in the data.

\section{Appendix}

In this supplementary material we present additional empirical results to the main paper.


\subsection{Model Illustration}
\begin{figure}[b]
    \centering
    \includegraphics[width=0.7\linewidth]{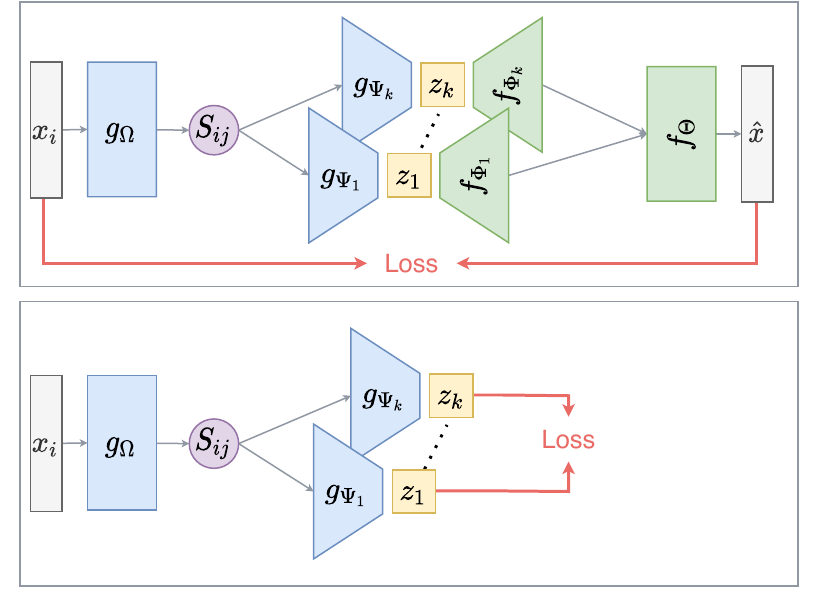}
    \caption{Model illustration.
    \textbf{Top:} reconstruction loss based partial tenzorised model.
    \textbf{Bottom:} tensorized model with loss on the obtained embeddings.
    }
    \label{fig:app_Illustraion}
\end{figure}

We now illustrate the two main tensorized models: firstly the reconstruction based model of the form
\begin{align*}
&\min_{{\{\Phi, \Psi\}^k_{1},\Theta,\Omega,S}}
\sum^n_{i=1}  \sum^k_{j = 1}S_{j,i}\left[\norm{\tilde{x}_{i,j} - \hat{x}_{i,j}
}^2 
\right]\nonumber\\
&\text{s.t. ~ } 1_k^TS = 1_n^T,\ S_{j,i} \geq 0,
\end{align*}
with $\hat{x}_{i,j}:=f_{\Theta}(f_{\Phi_j}(g_{\Psi_j}(g_{\Omega}(\tilde{x}_{i,j}))))$.
This general idea is used in the TAE (3) and TVAE (4) and illustrated in Figure~\ref{fig:app_Illustraion} (top).
Secondly the setting of partial tensorization that we defined as
\begin{align*}
&\min_{\{\Psi\}_{1}^k,\Omega,S}\frac{1}{n} \sum_{j=1}^n \sum_{i=1}^k S_{j,i}\left[\mathcal{L}\left(g_{\Psi_j}\left(g_\Omega(x_i)\right)\right)\right]\\
    &\text{s.t. ~ } 1_k^TS = 1_n^T,\ S_{j,i} \geq 0.\nonumber
\end{align*}
in (1) computed the loss on the obtained embeddings and build the foundation for the SSL setting. We illustrate this general idea in Figure~2 (bottom).

\subsection{TAE}
\begin{figure}[t]
    \centering
    \includegraphics[width=0.7\linewidth]{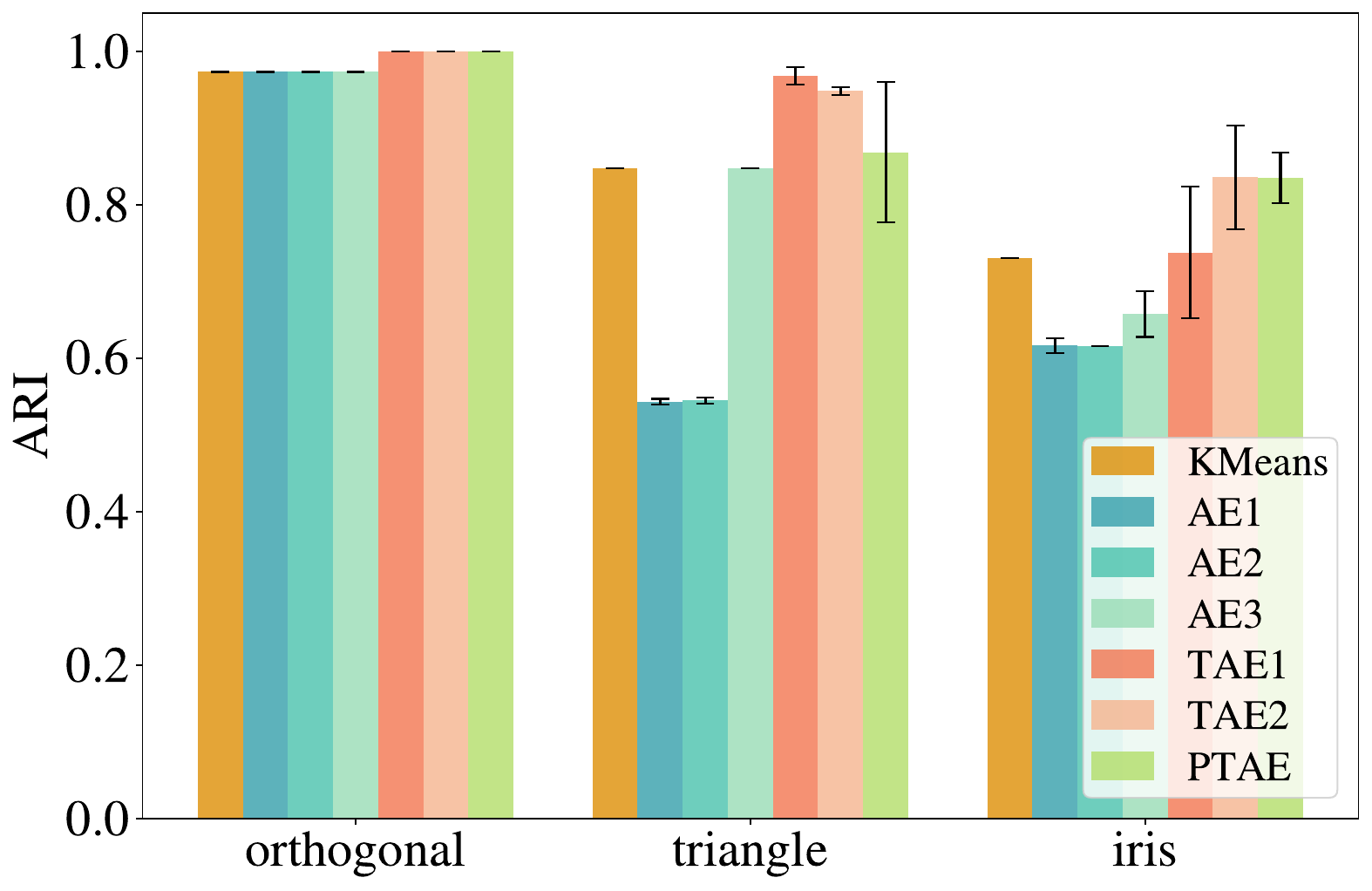}
    
    \includegraphics[width=0.7\linewidth]{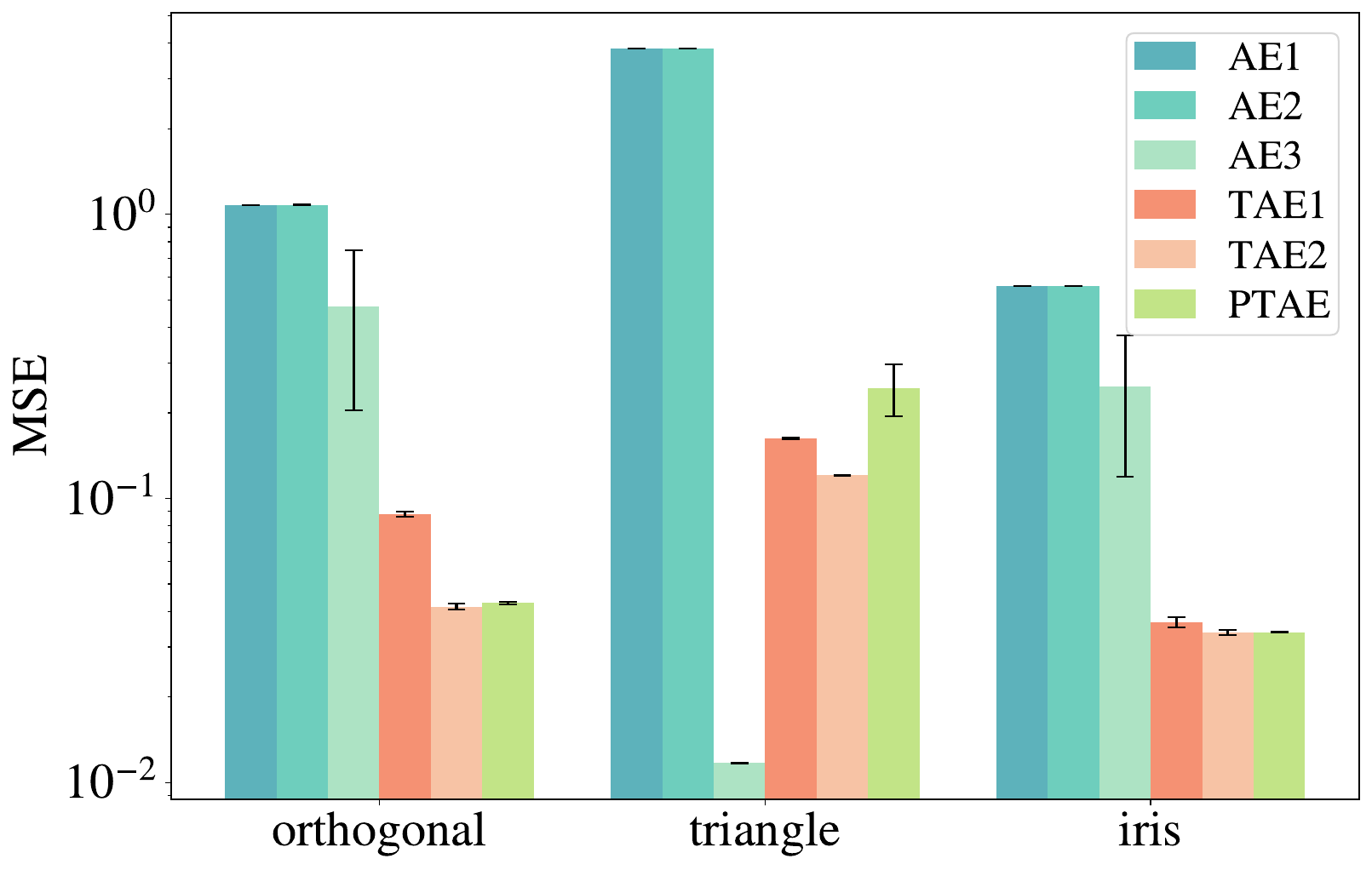}
    
    \includegraphics[width=0.7\linewidth]{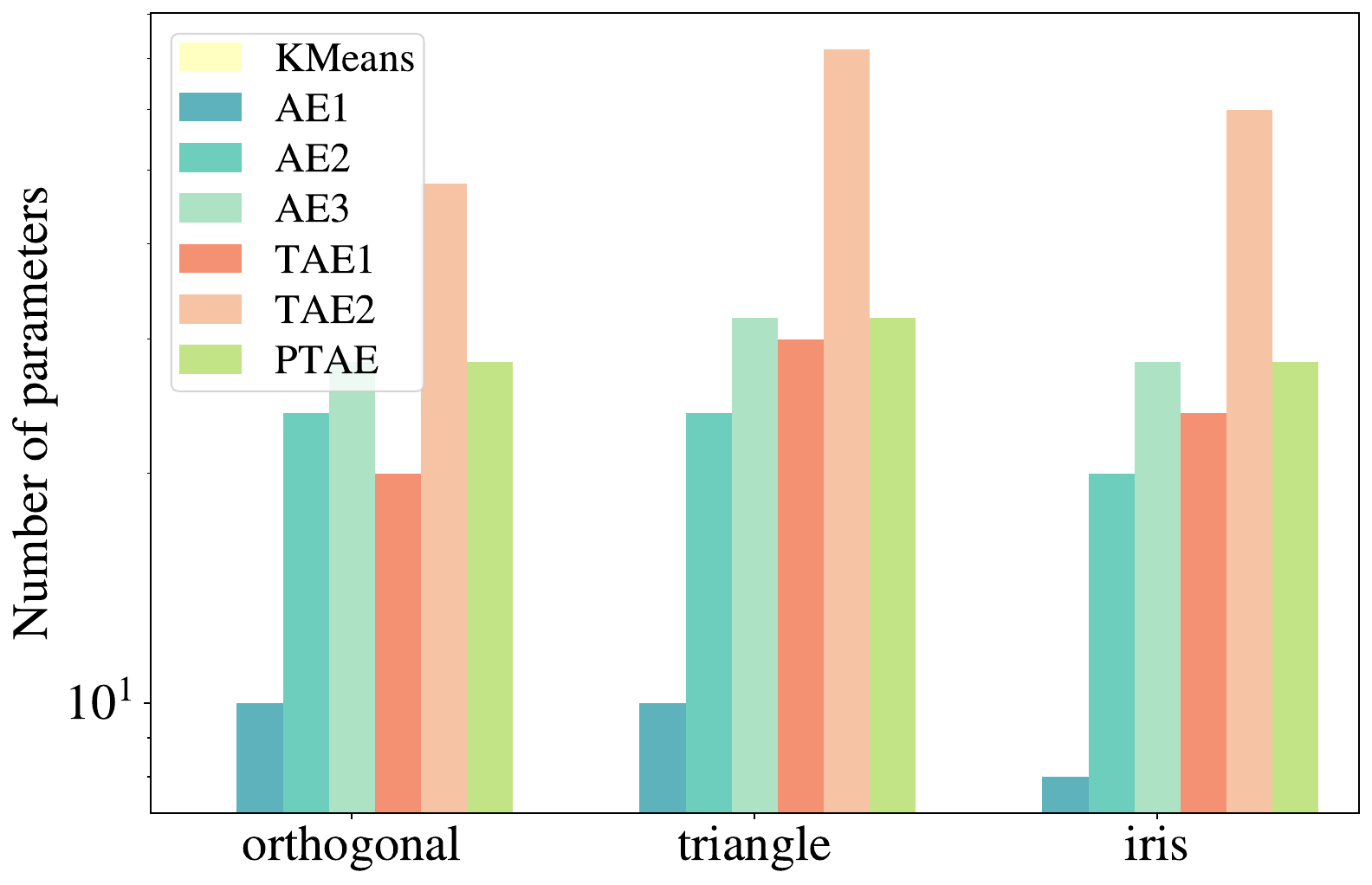}
    \caption{
    Additional datasets. Performance of AE, TAE and PTAE.
    \textbf{Top:} clustering obtained by different models with k-means as benchmark. Plotted is ARI (higher is better). 
    \textbf{Center:} de-noising of different models. MSE in log-scale (lower is better). 
    \textbf{Right:} number of parameters in each considered model.}
    \label{fig:TAE_app}
\end{figure}

\textbf{Additional Datasets.} Similar to Figure~2, we provide a similar analysis for additional datasets \emph{orthogonal} $(d=5,n=150,C=2)$ \cite{esser2023improved}, \emph{triangle} $(d=6,n=150,C=3)$ \cite{esser2023improved} and the real dataset \emph{iris} $(d=4,n=150,C=3)$ \cite{Fisher1936Iris1}.
We present the results in Figure~\ref{fig:TAE_app} and observe that the overall trends between AE, TAE and PTAE match the ones observed in the main section and Figure~2. Tensorization helps in learning better representation in all the cases.

\begin{figure}
    \centering
    \includegraphics[width=0.75\linewidth]{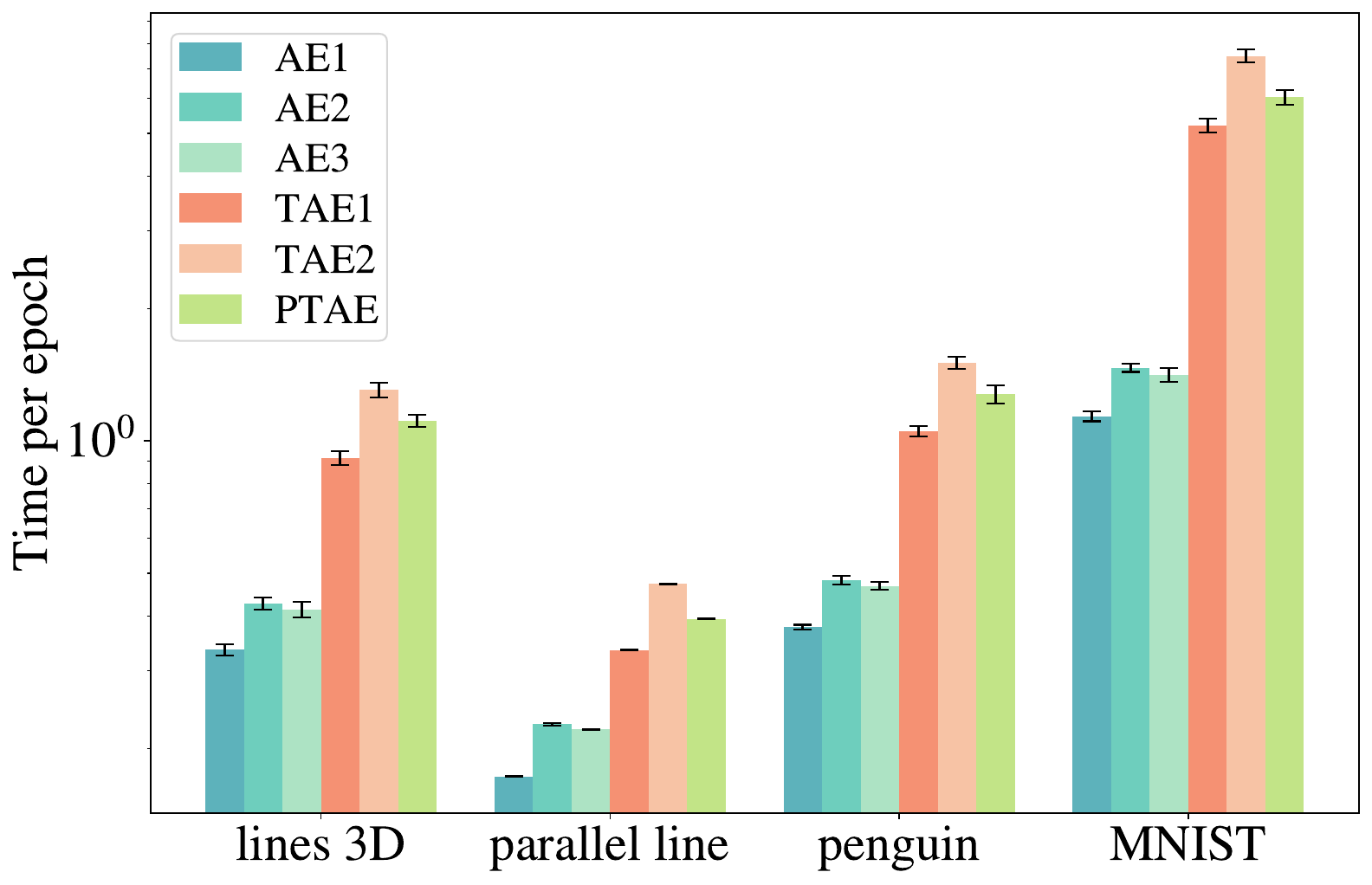}\\
    \hspace{-0.5cm}
    \includegraphics[width=0.8\linewidth]{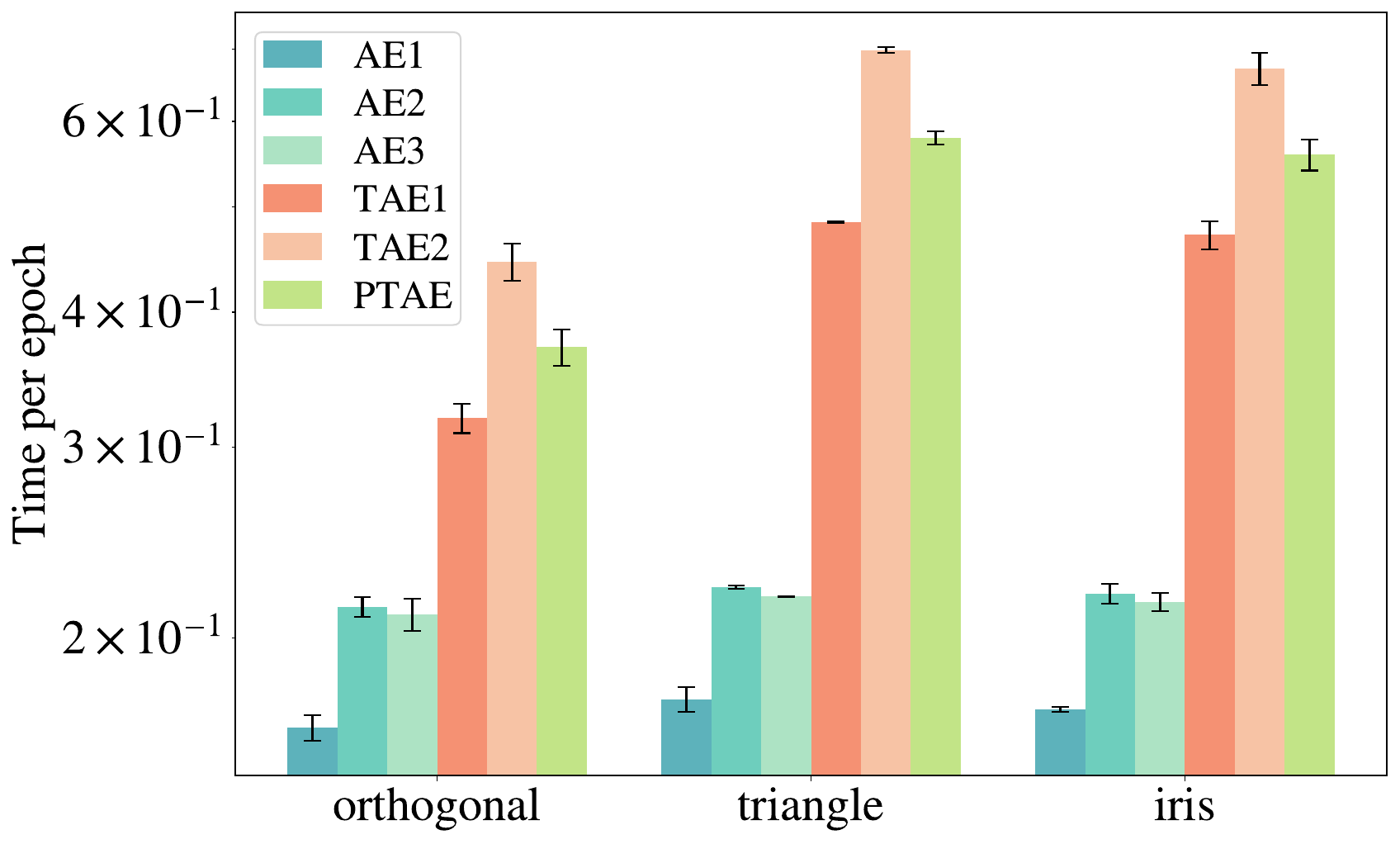}
    \caption{Runtime per epoch comparison between AE, TAE and PTAE.}
    \label{fig:ae_time}
\end{figure}

\textbf{Runtime.}
For the datasets considered in Figure~2 and Figure~\ref{fig:ae_time} (top row) we provide the runtime comparison in Figure~\ref{fig:ae_time} (bottom row). Plotted is the  run-time for one epoch average over five runs. 
Note that while TAE and PTAE have significantly higher runtime then standard AEs, this is due to a current naive implementation of the approach. The current implementation considers a sequential computation of the forward passes through the cluster specific encoding, making the approach slower. However this could be speed up significantly by parallel computing of the cluster specific embeddings.

\subsection{VAE}

\begin{figure}[t]
    \centering
    \includegraphics[width=0.24\linewidth]{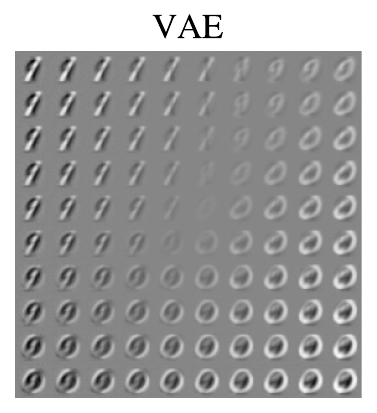}
    \includegraphics[width=0.24\linewidth]{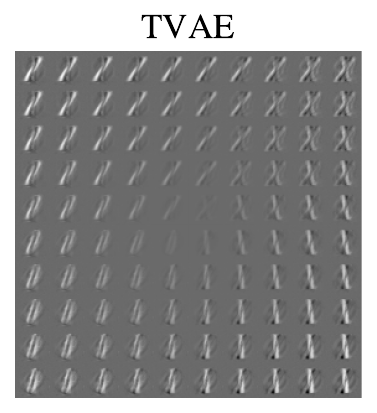}
    \includegraphics[width=0.24\linewidth]{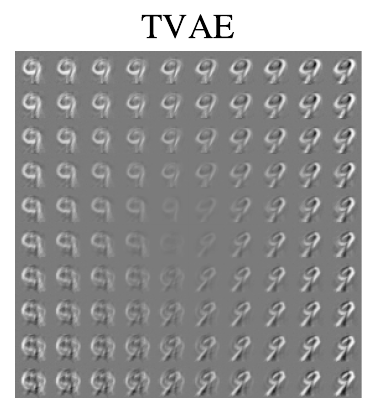}
    \includegraphics[width=0.24\linewidth]{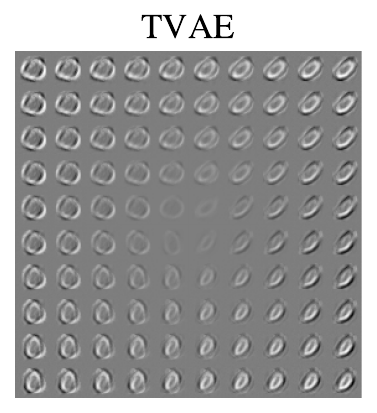}\\
    \includegraphics[width=0.24\linewidth]{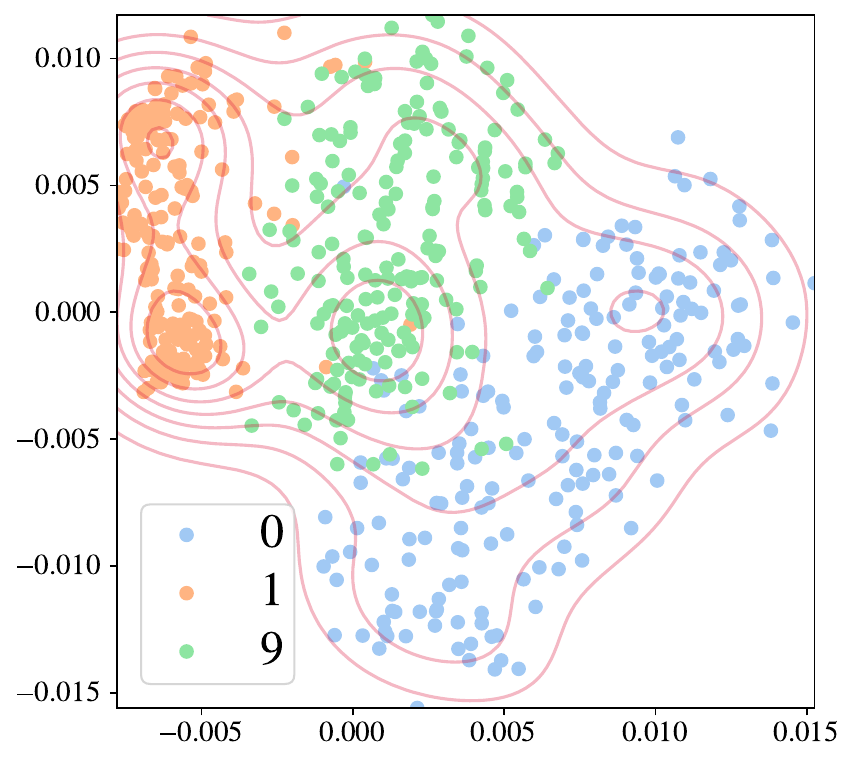}
    \includegraphics[width=0.24\linewidth]{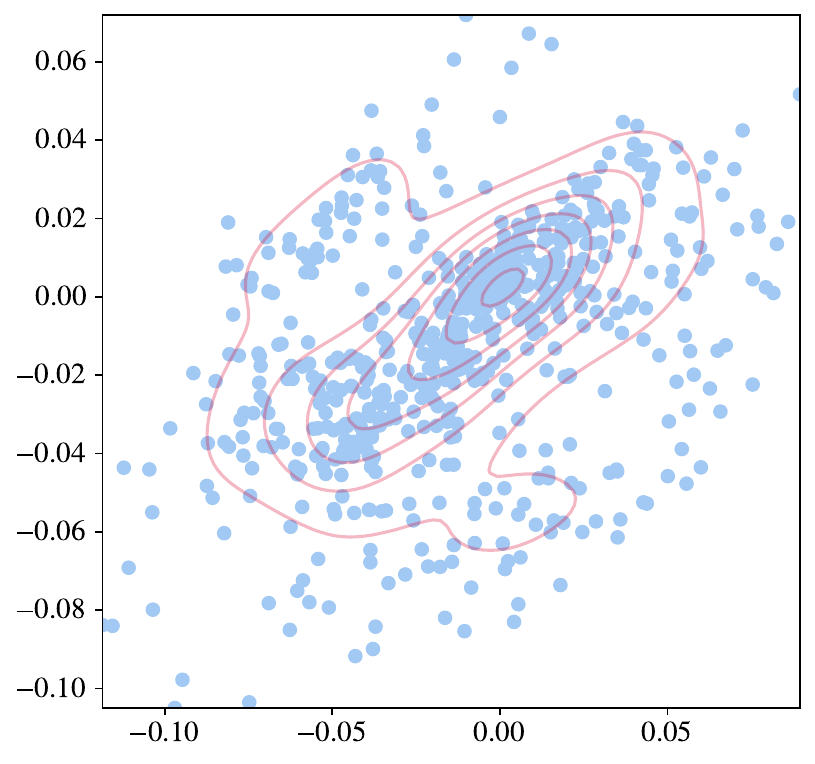}
    \includegraphics[width=0.24\linewidth]{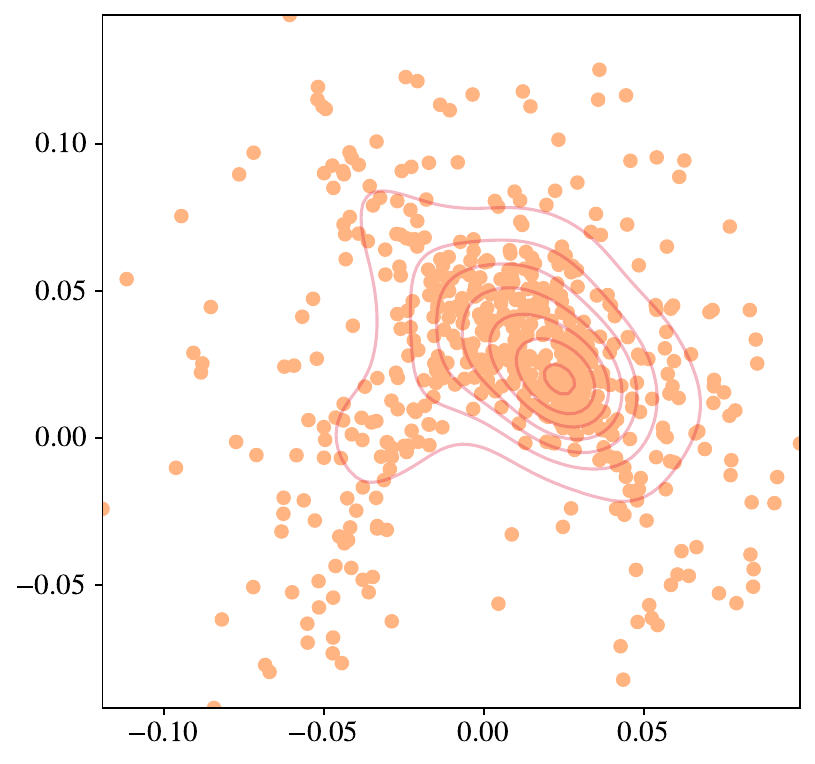}
    \includegraphics[width=0.24\linewidth]{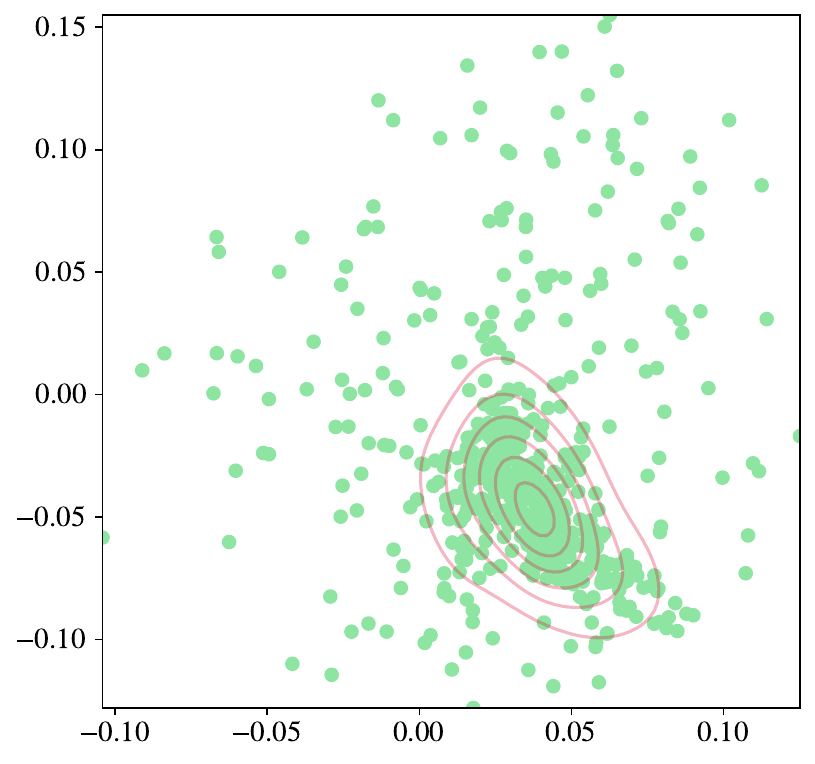}
    \caption{
    Same setup as Figure~3, however for
    MNIST with linear output layer and MSE loss.
     }
    \label{fig:vae_latent_space_MNIST_MSE}
\end{figure}

\textbf{Reconstruction Loss.}
In the main paper we defined the reconstruction loss through firstly applying a sigmoid function to the decoder output and then computing the loss using BCE. However this is not a unique choice for the optimization problem. One alternative formulation if using a linear output-layer and a MSE, such that the objective becomes:
    $\frac{1}{2n}\sum_i^n\sum_j^k S_{i,j}\norm{\tilde{x}_{i,j}-\hat{\mu}_{i,j}}^2.$
The results are presented in Figure~\ref{fig:vae_latent_space_MNIST_MSE}.

\begin{figure}[t]
    \centering
    \includegraphics[width=0.24\linewidth]{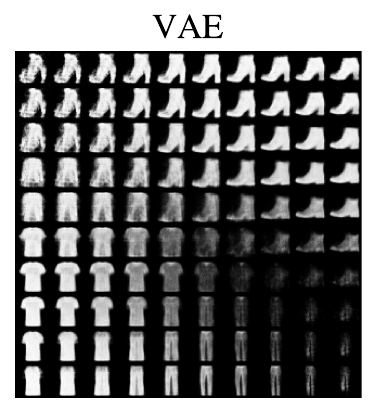}
    \includegraphics[width=0.24\linewidth]{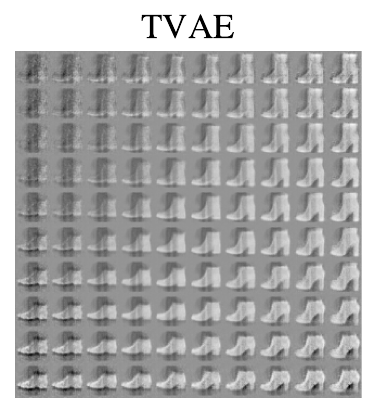}
    \includegraphics[width=0.24\linewidth]{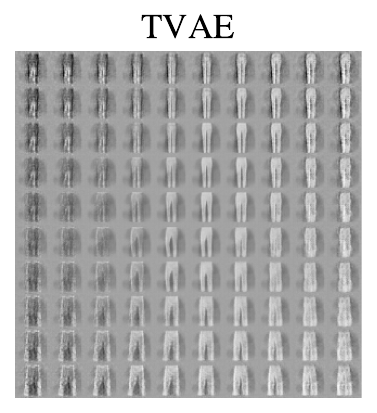}
    \includegraphics[width=0.24\linewidth]{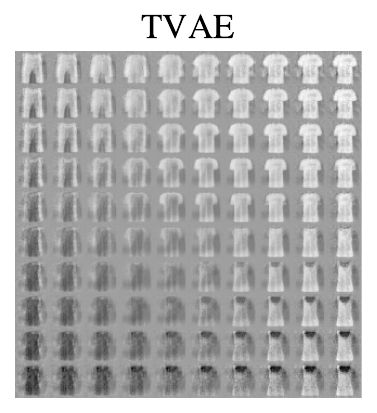}\\
    \includegraphics[width=0.24\linewidth]{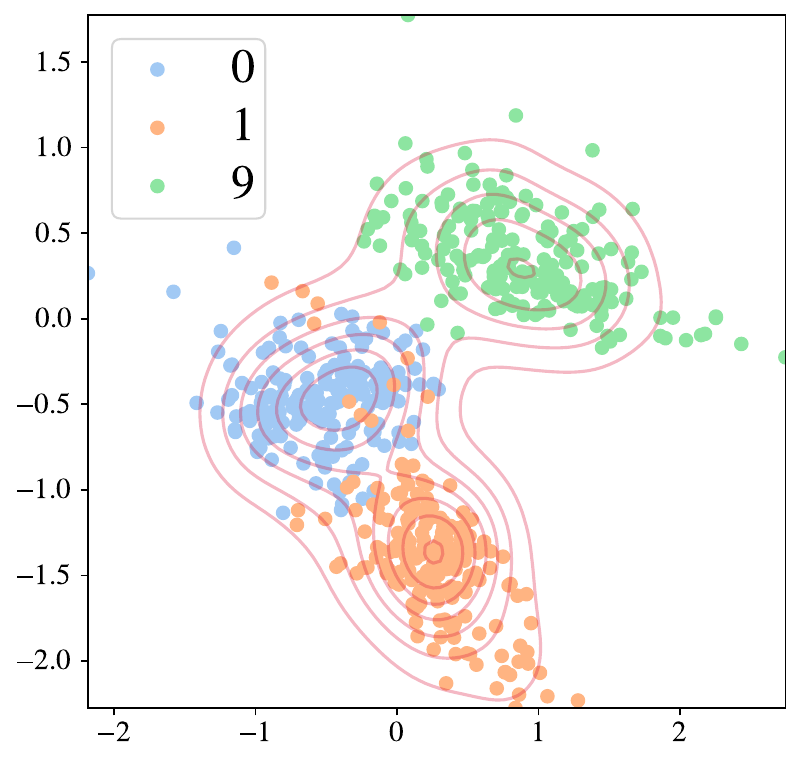}
    \includegraphics[width=0.24\linewidth]{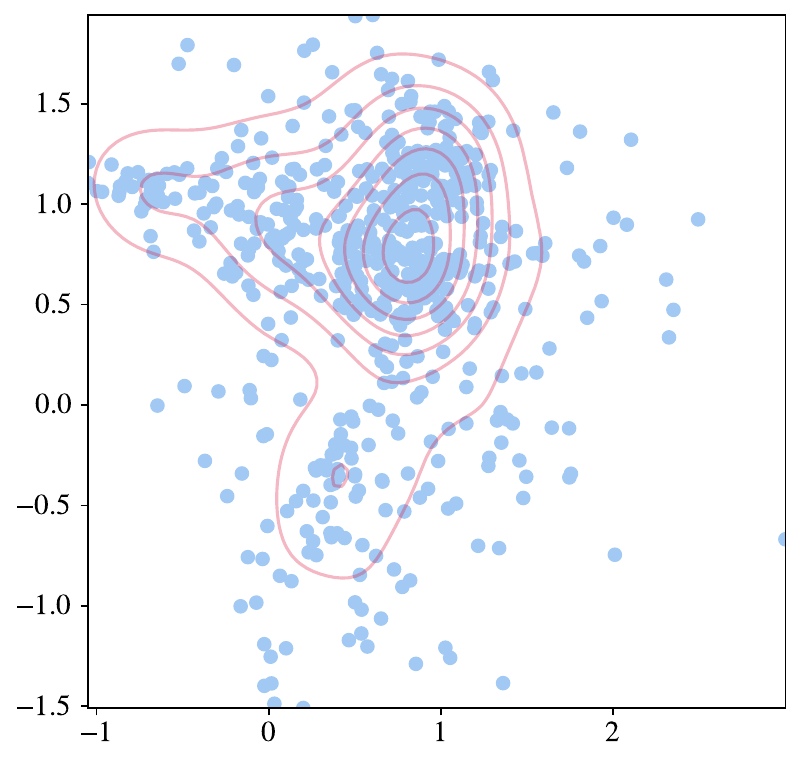}
    \includegraphics[width=0.24\linewidth]{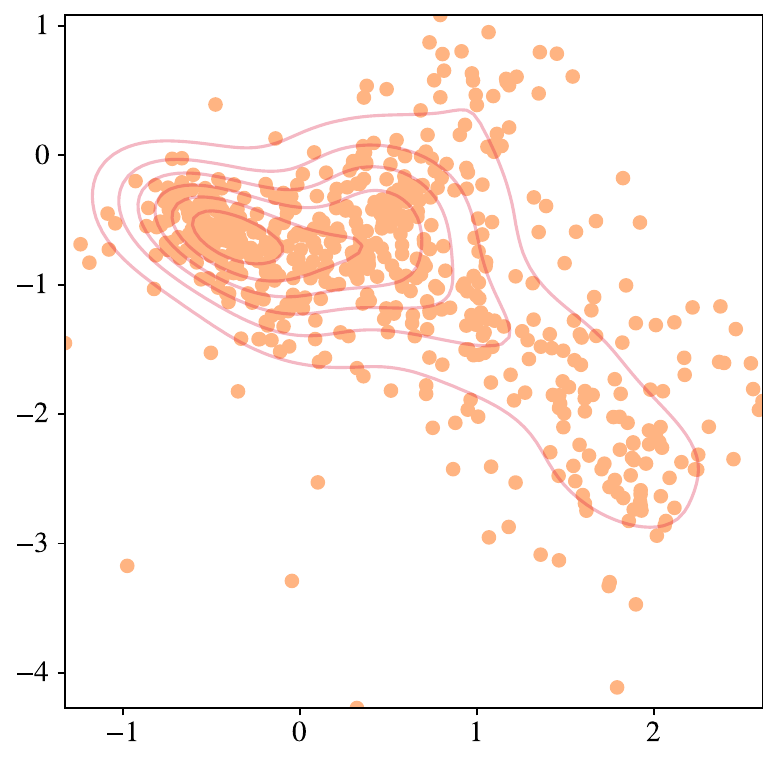}
    \includegraphics[width=0.24\linewidth]{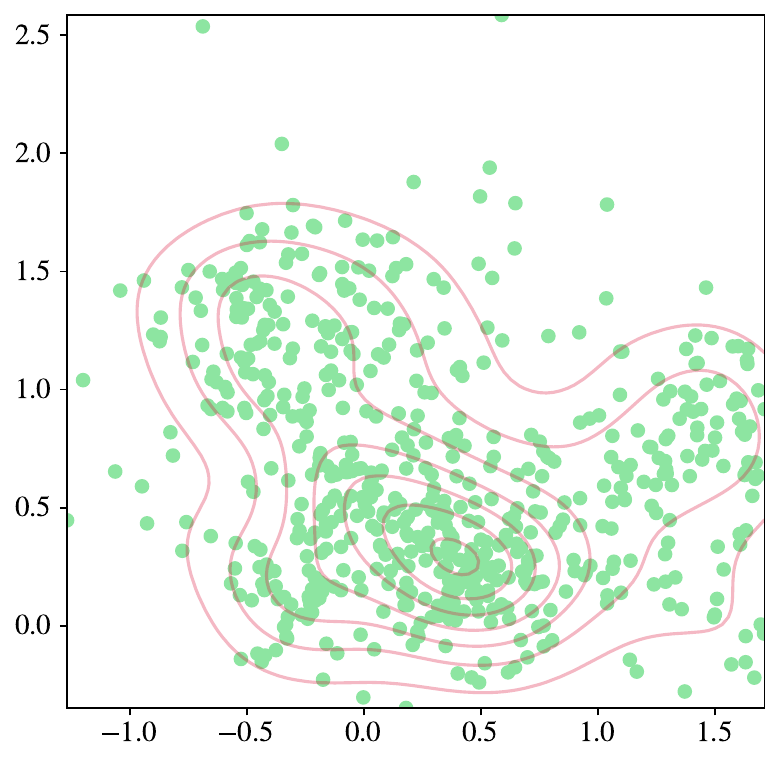}
    \caption{
        Same setup as Figure~3, however for
    FMNIST with BCE loss and CNN encoder an decoder.
    }
    \label{fig:vae_latent_space_fMNIST_bce}
\end{figure}

\textbf{Further datasets.}
In the main part of the paper we presented the results for MNIST. In Figure~\ref{fig:vae_latent_space_fMNIST_bce} we show corresponding results for Fashion MNIST \cite{FMNIST}.

\begin{figure}[t]
    \centering
    \includegraphics[width=0.24\linewidth]{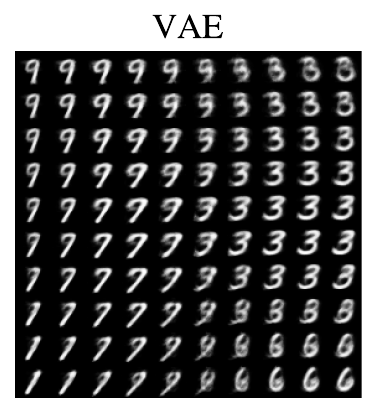}
    \includegraphics[width=0.24\linewidth]{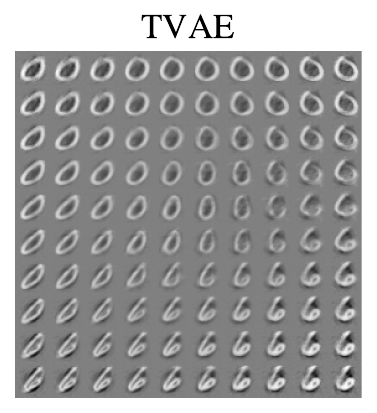}
    \includegraphics[width=0.24\linewidth]{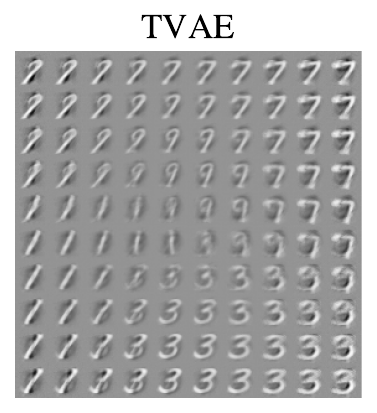}\\
    \includegraphics[width=0.24\linewidth]{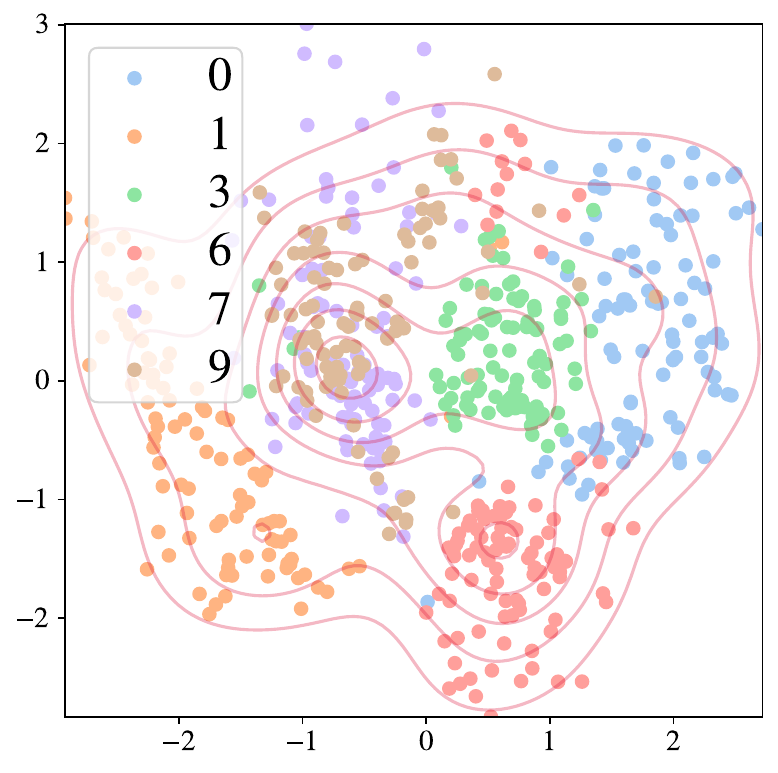}
    \includegraphics[width=0.24\linewidth]{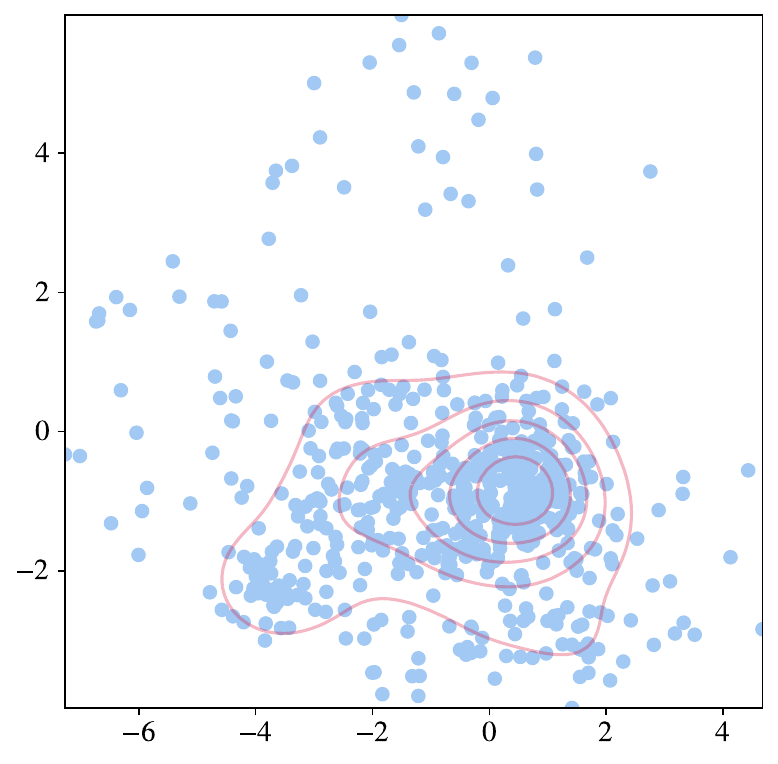}
    \includegraphics[width=0.24\linewidth]{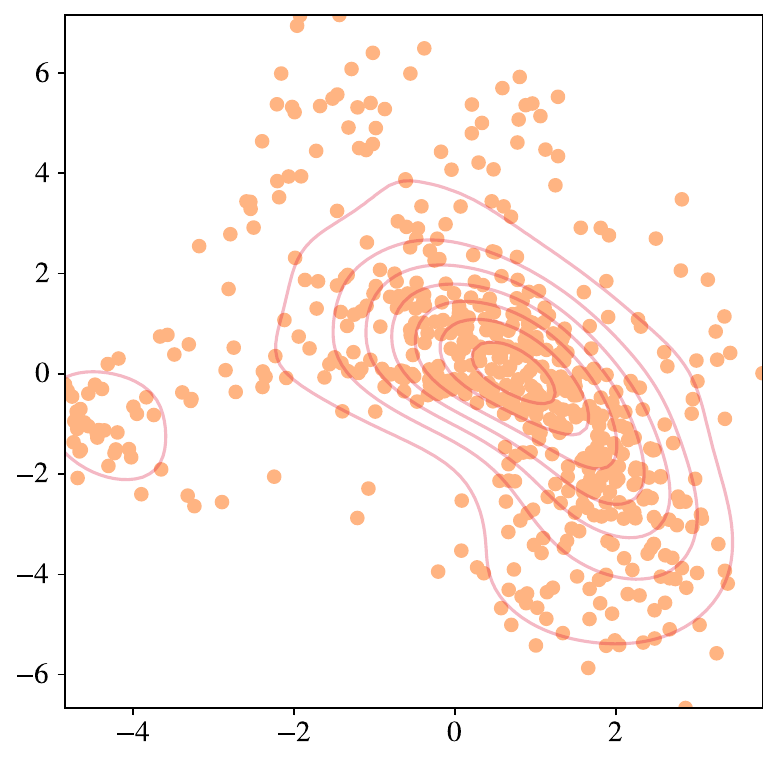}
    \caption{
        Same setup as Figure~3, however for
    MNIST with sigmoid output and BCE loss.
    The data set consists now of six classes $\{0,1,3,6,7,9\}$, however we only allow for two cluster specific encodings.
     }
    \label{fig:vae_latent_space_lessclust}
\end{figure}

\textbf{Number of considered clusters.} 
We illustrate the effect of underestimating the number of clusters by considering six classes of MNIST $\{0,1,3,6,7,9\}$ and learning only two cluster specfic embeddings using TVAE. The results are presented in Figure~\ref{fig:vae_latent_space_lessclust} showing the representations of similar classes are learned together where $\{0,6\}$ is learned in one and $\{1,3,7,9\}$ in the other. 

\subsection{SSL}

    \begin{figure}[h]
    \centering
    \includegraphics[width=0.6\linewidth]{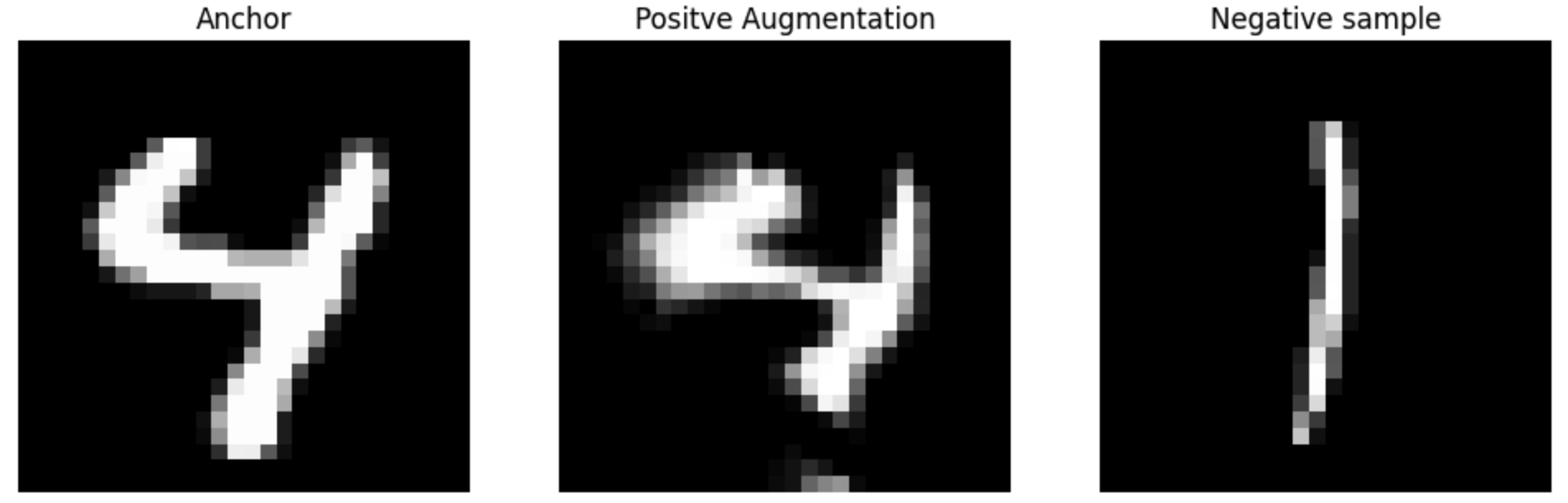}
    \caption{Illustration of the samples and elastic transformation applied on the MNIST dataset. 
    \textbf{Left:} anchor image.
    \textbf{Middle:} augmented version of the anchor image through elastic transformation.
    \textbf{Right:} negative, sample (independent sampled from the data-set).
    }
    \label{fig:Augmentations}
\end{figure} 

\textbf{Elastic Transformation.}  We now illustrate the data augmentation we consider for the SSL experiments.
    The elastic transformation function is used to create augmented positive samples by deforming the given anchor image. The function applies random, smooth distortions to the image by generating displacement fields using Gaussian filters. This perfectly simulates the different styles of writing found in the MNIST dataset as illustrated in Figure~\ref{fig:Augmentations}.

\subsection{RBM}

\begin{figure}[h]
    \centering
    \includegraphics[width=0.24\linewidth]{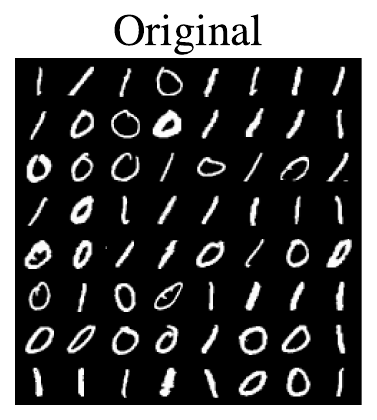}
    \includegraphics[width=0.24\linewidth]{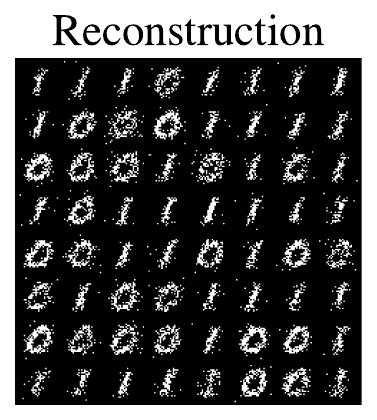}
    \caption{Reconstruction using a standard RBM. 
    \textbf{Left:} original image.
    \textbf{Right:} reconstruction.
    }
    \label{fig:rbm_no_tensor}
\end{figure}

\textbf{Reconstruction of standard RBM.} In the main paper we presented the cluster specific reconstruction using the TRBM model in Figure~5. For comparison we furthermore consider the reconstruction obtained by the standard RBM as show in Figure~\ref{fig:rbm_no_tensor}.

\section*{Acknowledgments}
This work was done when Omar Al-Dabooni was a Bachelor’s student at the Technical University of Munich. The authors acknowledge the funding assistance from projects, specifically the German Research Foundation (Priority Program SPP 2298, project GH 257/2-1) and the TUM Georg Nemetschek Institute Artificial Intelligence for the Built World.

\bibliographystyle{IEEEtran}
\bibliography{main}


 


\vspace{11pt}


\vfill


\end{document}